\documentclass[10pt,twocolumn,letterpaper]{article}

\usepackage{iccv}              
\usepackage[accsupp]{axessibility}

\definecolor{iccvblue}{rgb}{0.21,0.49,0.74}
\usepackage[pagebackref,breaklinks,colorlinks,allcolors=iccvblue]{hyperref}

\def\paperID{} 
\def\confName{}
\def\confYear{}

\title{\ourmethod: Part-Guided Texturing for 3D Human Reconstruction \\ from a Single Image}

\author{
  Hyeongjin Nam$^{*1}$ \hskip1.6em Donghwan Kim$^{*1}$ \hskip1.6em Gyeongsik Moon$^{\dag2}$ \hskip1.6em Kyoung Mu Lee$^{1}$ \vspace{+2mm} \\ 
   $^{1}$Dept. of ECE\&ASRI, Seoul National University \hskip1.0em $^{2}$Dept. of CSE, Korea University  \\
   {\tt\small \{namhjsnu28, dh971106, kyoungmu\}@snu.ac.kr, mks0601@korea.ac.kr} \\
   \small{\url{https://hygenie1228.github.io/PARTE/}}
}

\usepackage[dvipsnames]{xcolor}

\usepackage{graphicx}
\usepackage{amsmath}
\usepackage{amssymb}
\usepackage{booktabs}

\usepackage{times}
\usepackage{epsfig}

\usepackage{verbatim}

\usepackage{algorithm}
\usepackage{algorithmicx}
\usepackage{algpseudocode}

\usepackage{color}

\usepackage{ctable}
\usepackage{makecell}
\usepackage{tabularx}
\usepackage{multirow}
\usepackage{multicol}
\usepackage{arydshln}

\usepackage{pifont}
\usepackage{cuted}
\newcommand{\ourmethod}{PARTE}
\renewcommand{\thefootnote}{\fnsymbol{footnote}}

\begin{document}
\maketitle
\renewcommand{\thefootnote}{} 
\footnotetext[1]{\hspace*{-1.0em}\textsuperscript{*}Equal contribution. \hskip1.0em  \textsuperscript{\dag}Corresponding author.}
\footnotetext[2]{\hspace*{-1.0em}\textsuperscript{\ddag}This work was done while Hyeongjin Nam was in KRAFTON.}

\begin{abstract}
The misaligned human texture across different human parts is one of the main limitations of existing 3D human reconstruction methods.
Each human part, such as a jacket or pants, should maintain a distinct texture without blending into others.
The structural coherence of human parts serves as a crucial cue to infer human textures in the invisible regions of a single image.
However, most existing 3D human reconstruction methods do not explicitly exploit such part segmentation priors, leading to misaligned textures in their reconstructions.
In this regard, we present \textbf{\ourmethod}, which utilizes 3D human part information as a key guide to reconstruct 3D human textures. 
Our framework comprises two core components. 
First, to infer 3D human part information from a single image, we propose a 3D part segmentation module (PartSegmenter) that initially reconstructs a textureless human surface and predicts human part labels based on the textureless surface. 
Second, to incorporate part information into texture reconstruction, we introduce a part-guided texturing module (PartTexturer), which acquires prior knowledge from a pre-trained image generation network on texture alignment of human parts.
Extensive experiments demonstrate that our framework achieves state-of-the-art quality in 3D human reconstruction.
\end{abstract}
\section{Introduction}
\begin{figure}[t!]
  \centering
\includegraphics[width=1.0\linewidth]{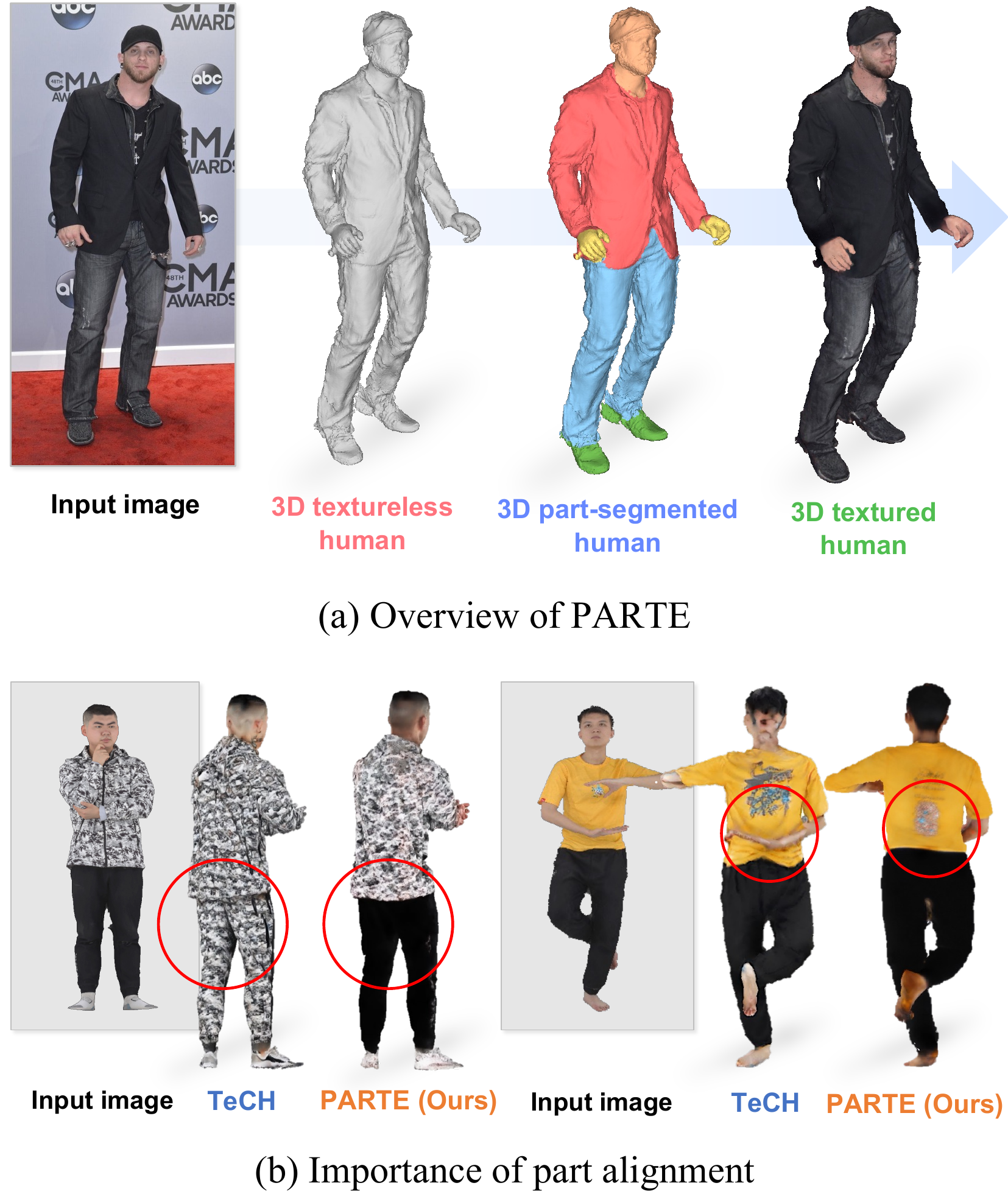}
  \vspace{-1.4em}
  \caption{\textbf{Overview of \ourmethod.}
  Our framework reconstructs a realistic 3D textured human from a single image by effectively improving part alignment with 3D human part information.
  }
  \vspace{-0.8em}
  \label{fig:teaser}
\end{figure}

3D human reconstruction aims to recover both the geometry and texture of a human from a single image.
This task is essential for various applications, such as filmmaking, game development, and AR/VR.
The key challenge in 3D human reconstruction is inferring human textures for invisible regions by extracting appearance cues, such as cloth coverage and hair color, from the limited front-view image.

Despite remarkable advances in 3D reconstruction of human geometry~\cite{men2024en3d,xiu2023econ,xiu2022icon,saito2020pifuhd,han20232k2k}, reconstructing high-fidelity textures has been relatively unexplored.
Most existing methods~\cite{saito2019pifu,zheng2020pamir,ho2024sith,chen2024generalizable,alldieck2022photorealistic,zhang2023globalcorrelated,huang2020arch,he2021arch++,xue2024human3diff} train their networks on 3D human scan datasets~\cite{renderpeople,tao2021function4d,cai2022humman,ma2020learning,ho2023learning} to directly predict the colors corresponding to 3D points on the human surface.
More recently, several works~\cite{huang2024tech,zhang2024humanref,albahar2023humansgd,zhang2024sifu,zhan2024shert,kolotouros2024avatarpopup} have leveraged a pre-trained diffusion model~\cite{rombach2022ldm} that contains rich image prior knowledge to reconstruct realistic human textures.
However, a critical challenge remains in human texture reconstruction: part misalignment. 
Existing methods often fail to reconstruct textures that accurately correspond to each human part.
They primarily rely on the global context of the image rather than explicitly identifying local semantics, which makes it difficult to capture part-specific information from the input image.
Furthermore, existing methods often ignore the structural information of the human, resulting in incorrect textures that do not match the underlying structure.
As shown in \cref{fig:teaser} (b), these failure cases highlight the requirement for part alignment to enhance the overall texture reconstruction quality.

\begin{figure*}[t!]
  \centering
  \includegraphics[width=1.0\linewidth]{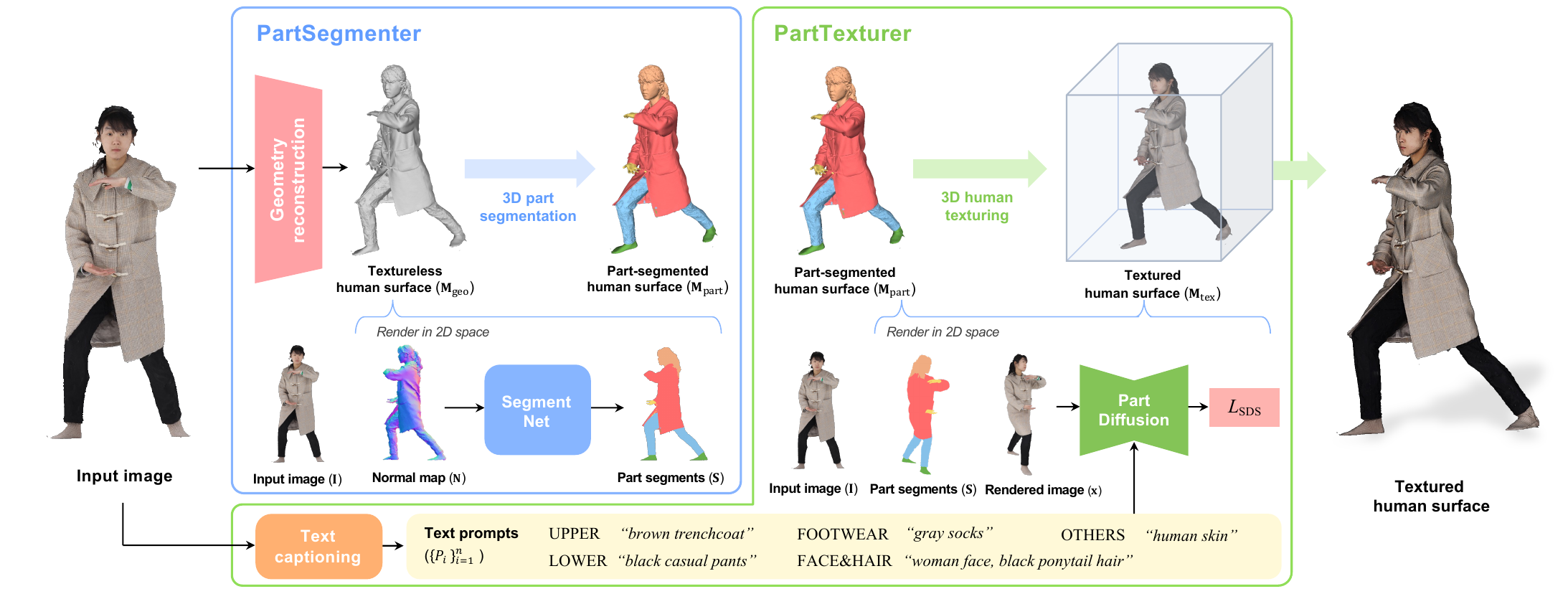}
  \vspace{-1.4em}
  \caption{\textbf{Overall pipeline of \ourmethod.}
  Our framework initially reconstructs a textureless human mesh from a single image and textures it based on two core modules.
  \textcolor[RGB]{70,110,255}{\textbf{PartSegmenter}} predicts 3D human parts of the textureless human mesh by incorporating information from the input image and normal maps of the textureless human surface.
  Based on the 3D human part segmentations, \textcolor[RGB]{50,190,50}{\textbf{PartTexturer}} reconstructs the human textures using PartDiffusion network that infers plausible human appearance corresponding to the human parts.
  }
  \vspace{-0.2em}
  \label{fig:pipeline}
\end{figure*}

To address the part misalignment, we propose \textbf{\ourmethod}, \textbf{PAR}t-guided \textbf{TE}xturing for 3D human reconstruction.
This framework estimates 3D part segmentations of the human surface (\textit{e.g.}, upper-cloth and footwear) and utilizes them as main guidance for reconstructing human textures, as shown in \cref{fig:pipeline}.
The part segmentations effectively guide subsequent texturing by specifying distinct human part regions, resulting in well-aligned human textures corresponding to the human parts. 
Our framework consists of two components: 1) \textcolor[RGB]{70,110,255}{\textbf{PartSegmenter}} and 2) \textcolor[RGB]{50,190,50}{\textbf{PartTexturer}}.
\textcolor[RGB]{70,110,255}{\textbf{PartSegmenter}} is a module that predicts 3D human part segmentations from an input image.
The key challenge of this module lies in predicting part labels for regions not visible in the front-view. 
To address this, PartSegmenter initially reconstructs the textureless 3D human surface and infers the part segmentation based on the textureless surface.
The textureless human surface contains structural cues, such as depth variations and human part boundaries, which cannot be captured in the front-view, aiding inference of part labels for invisible regions.
By incorporating such geometric cues with the visual feature of the image, PartSegmenter effectively resolves ambiguity in invisible regions, enabling accurate and consistent 3D part segmentation.

\textcolor[RGB]{50,190,50}{\textbf{PartTexturer}} is a module that reconstructs human textures based on our estimated 3D part segmentations as key guidance. 
The key challenge of this module is to infer a realistic human texture that satisfies two constraints: alignment with the segmented human parts and visual consistency with the input image.
To this end, we devise a specialized diffusion network, which incorporates both the part segmentations and the visual features extracted from the input image.
From the part segmentations, the model receives structural guidance, ensuring that textures are assigned accurately to distinct part regions with clear boundaries.
From the input image, it captures detailed visual features of each human part, providing exact texture guidance for each respective part.
By leveraging the diffusion network’s effectiveness in part-aware texturing, our framework accurately reconstructs human textures corresponding to the part segmentations, thus improving the overall quality and consistency of the reconstructed human textures.

As a result, our proposed \ourmethod~significantly improves the visual fidelity of reconstructed 3D humans, addressing part misalignment issues, compared to existing state-of-the-art 3D human reconstruction methods.
Our contributions can be summarized as follows.
\begin{itemize}
\item We propose \textbf{\ourmethod}, a 3D human reconstruction framework that recovers realistic human textures with precise alignment to each human part.
\item To acquire precise 3D part segmentation, we present a part segmentation module, PartSegmenter, which integrates information from both visible and invisible regions of the input image.
\item To accurately reconstruct human textures that align with human parts, we propose a part-guided texturing module, PartTexturer, which effectively combines part segmentations and visual features from the input image.
\end{itemize}

\section{Related Works}
\noindent\textbf{3D human reconstruction.}
3D human reconstruction methods~\cite{weng2019photo,lazova2019360,alldieck2019learning} can be categorized into regression- and optimization-based approaches.
The regression-based methods~\cite{saito2019pifu,saito2020pifuhd,zheng2020pamir,alldieck2022phorhum,huang2020arch,he2021arch++,corona2023structured,liao2023high,wang2023complete,peng2024charactergen,ho2024sith,pan2025humansplat,moon20223d,nam2025decloth} directly estimate geometry (\textit{i.e.}, distance fields) and texture (\textit{i.e.}, RGB color) from an input image, by training neural networks on 3D human scan datasets.
PIFu~\cite{saito2019pifu} is proposed to spatially align image features with the input image to learn geometry and texture effectively.
SiTH~\cite{ho2024sith} adopts a two-stage approach, first predicting a back-view image, then reconstructing a 3D human using both the front- and back-view images.
However, such approaches are fundamentally limited by the scarcity of 3D training data, as acquiring high-quality 3D human scans requires specialized studios or a process of capturing multi-view images.
To address this limitation, more recent methods~\cite{sengupta2024diffhuman,huang2024tech,ho2024sith,zhan2024shert,wang2024geneman,gao2024contexhuman,zhang2024sifu} have adopted an optimization-based approach that takes advantage of the prior knowledge of pre-trained diffusion models~\cite{rombach2022ldm} on large-scale 2D image datasets~\cite{fu2022stylegan,lin2014microsoft,liu2016deepfashion}.
Since large-scale 2D images are relatively easy to collect, the optimization-based approach has demonstrated realistic 3D reconstructions~\cite{jain2022dreamfield,poole2022dreamfusion,chen2023fantasia3d,wang2023prolificdreamer}.
TeCH~\cite{huang2024tech} utilizes a text-to-image diffusion model for reconstruction, ensuring the appearance of the reconstructed 3D human matches the text information of the image.
SIFU~\cite{zhang2024sifu} adopts a coarse-to-fine manner, first reconstructing human texture through a regression-based method and refining it via an optimization-based method.

Recently, several works~\cite{xu2021texformer,han20232k2k,zhang2024humanref} have utilized human part priors in their 3D human reconstruction pipelines.
Texformer~\cite{xu2021texformer} and 2K2K~\cite{han20232k2k} use 2D part segmentations from the input image without explicitly considering the overall structure of 3D human parts.
HumanRef~\cite{zhang2024humanref} uses hand-crafted 3D bounding boxes of human parts in the reconstruction, but the bounding boxes do not reflect the actual human parts in the image.
Unlike these methods, our proposed~\ourmethod~explicitly infers dense 3D part labels in 3D space, providing two key advantages.
First, since our framework estimates part segmentations in 3D space, the 3D part segmentations can provide accurate part guidance from any viewpoint instead of relying solely on front-view information. 
Second, our estimated 3D part segmentations reflect the actual appearance of the input image, ensuring a correct texture assignment to each human part.
With these advantages, our \ourmethod~effectively reconstructs human textures with precise alignment across human parts.

\noindent\textbf{3D human generation.}
Alongside image-based 3D human reconstruction methods, 3D human generation~\cite{grigorev2021stylepeople,alldieck2021imghum,dong2023ag3d,hong2023eva3d,xiong2023get3dhuman,chen2023primdiffusion,xu2023xagen,hu2024structldm, huang2024humannorm,men2024en3d,fu2023text,hong2022avatarclip,kim2023chupa,jiang2023avatarcraft,huang2024dreamwaltz,cao2024dreamavatar,kolotouros2023dreamhuman,liao2024tada,kolotouros2024avatarpopup,abdal2024gsm} using different modalities (\textit{e.g.}, text prompt) has also been actively studied.
AG3D~\cite{dong2023ag3d} trains a 3D human generation network with adversarial loss to learn realistic human appearances.
AvatarCLIP~\cite{hong2022avatarclip} proposes to generate 3D humans whose rendered images are conditioned to align with text descriptions using a pre-trained vision-language model~\cite{radford2021clip}.
Unlike 3D human generation, image-based 3D human reconstruction has the additional challenge of preserving the appearance of the input image, such as its human part structure.
To address this challenge, we propose part-guided texturing that effectively reconstructs human textures while preserving human appearance in the image.

\noindent\textbf{Region-aware image generation.}
Region-aware image generation~\cite{park2019semantic,li2019object,sun2019image,zhu2020sean,li2021collaging,bar2023multidiffusion,zheng2023layoutdiffusion,yang2023reco,wang2024instancediffusion,dong2018softgated,han2019clothflow,zhang20223dsgan,lv2021learning,song2019unsupervised} aims to synthesize images where each region is generated in alignment with its intended semantics, enabling more structured and controllable image generation.
LayoutDiffusion~\cite{zheng2023layoutdiffusion} modulates the diffusion network to generate images that conform to text prompts and 2D bounding boxes.
InstanceDiffusion~\cite{wang2024instancediffusion} improves region-aware control by conditioning generation on region-specific prompts within corresponding regions.
In 3D human reconstruction, the human body naturally consists of distinct regions, human parts, each with its own structural and visual significance.
Building on the strong generative power of region-aware image generation, we introduce part-guided texturing, which infers human textures explicitly guided by human part information.

\begin{figure}[t!]
  \centering
  \includegraphics[width=1.0\linewidth]{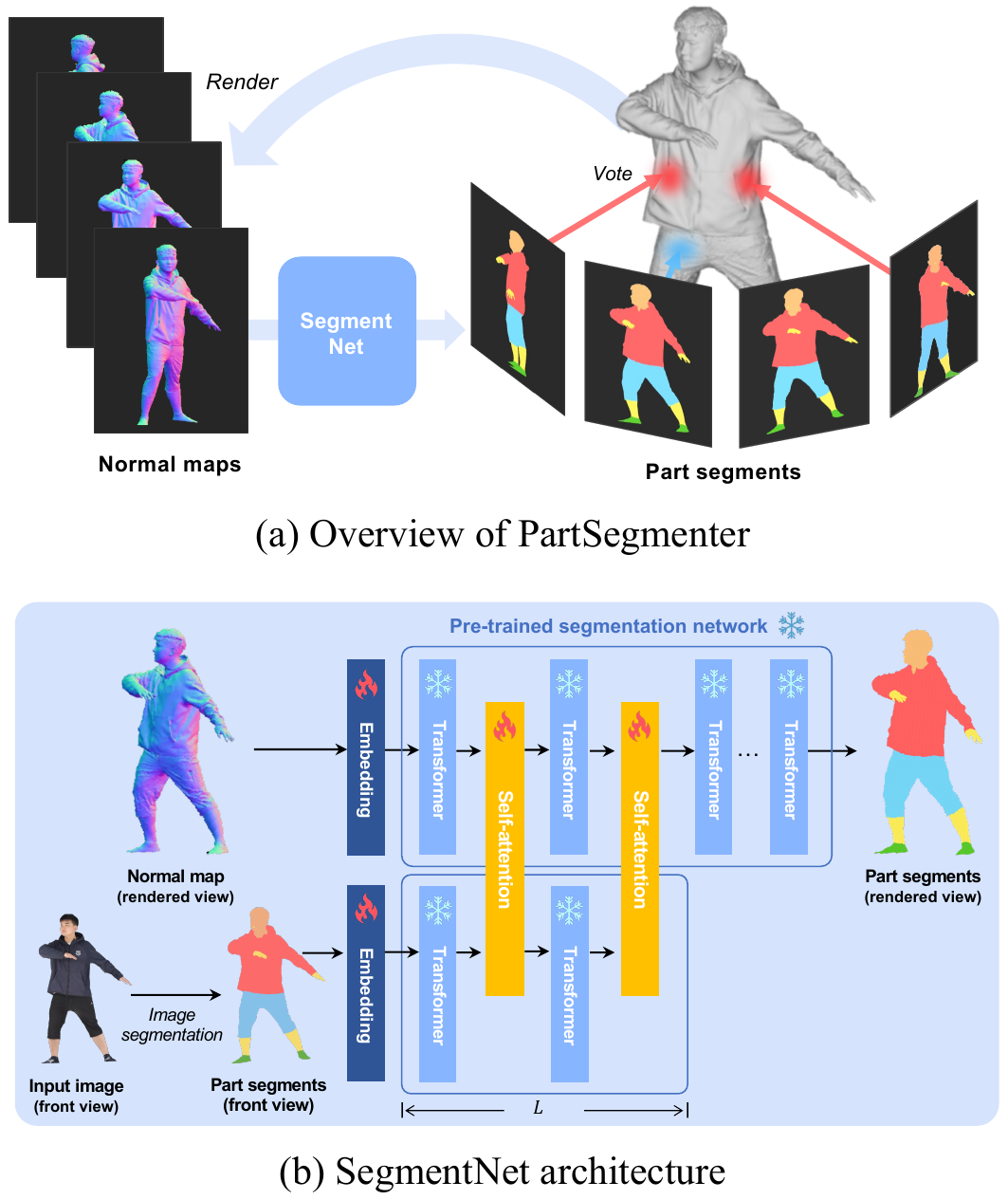}
  \vspace{-1.3em}
  \caption{\textbf{Detailed process of \textcolor[RGB]{70,110,255}{\textbf{PartSegmenter}}.}
    (a) From a textureless human surface, PartSegmenter performs normal rendering and acquires part segments via SegmentNet.
    The part segments are used to vote for 3D human part labels on the human surface.
    (b) SegmentNet incorporates features from the normal map and the front-view image for accurate part segmentation.
  }
  \vspace{-0.4em}
  \label{fig:normal-based_segmentor}
\end{figure}
\section{PartSegmenter}
\label{sec:partsegmenter}
\textcolor[RGB]{70,110,255}{\textbf{PartSegmenter}} estimates the 3D part-segmented human surface $\textbf{M}_{\text{part}}$ from an input image, where each vertex on the surface is assigned a semantic label of $n=5$ human parts: face \& hair, upper-clothes, lower-clothes, footwear, and others.
The acquired 3D part-segmented surface serves as input for the next module, PartTexturer (\cref{sec:PartTexturer}), which reconstructs human textures.

\subsection{Geometry reconstruction}
Given an input image, we first reconstruct the high-fidelity geometry $\textbf{M}_{\text{geo}}$ of a 3D human surface before predicting the 3D part-segmented human surface $\textbf{M}_{\text{part}}$.
For our geometry reconstruction, we employ an off-the-shelf reconstruction method, TeCH~\cite{huang2024tech} that has shown superior performance in real-world scenarios.
Note that our framework is compatible with various 3D human geometry reconstruction methods.
A detailed explanation of the geometry reconstruction process is provided in the supplementary material.

\subsection{3D part segmentation} 
\cref{fig:normal-based_segmentor} (a) illustrates the process of estimating the 3D segmented human surface $\textbf{M}_{\text{part}}$ from the initially reconstructed textureless human surface $\textbf{M}_{\text{geo}}$.
From the textureless human surface, we render a normal map $\textbf{N}^{\textbf{k}}$, where $\textbf{k}$ is a camera viewpoint sampled uniformly from a spherical distribution.
With the normal map, SegmentNet predicts its 2D part segments $\textbf{S}^{\textbf{k}}$, assigning one of $n$ part labels to each region.
We collect 2D part segments from 30 uniformly distributed viewpoints and aggregate them via voting to construct the part-segmented human surface $\textbf{M}_{\text{part}}$.

This strategy of aggregating 2D segmentation in 3D space, rather than directly predicting part labels in 3D, enhances generalization.
A naive approach to 3D part segmentation is to predict part segments directly in 3D space~\cite{qi2017pointnet,zhao2021point,qi2017pointnet++,takmaz23iccv} by training a prediction network on 3D human datasets.
However, these 3D datasets require restricted environments, such as capture studios, which limits their diversity and fails to capture real-world scenarios.
Consequently, designing a pipeline to directly predict 3D part labels results in poor generalization.
In contrast, 2D human datasets~\cite{fu2022stylegan,lin2014microsoft,liu2016deepfashion} are relatively abundant, enabling more generalizable 2D human part segmentation.
To leverage this advantage, we adopt a strategy that first performs segmentation in a 2D space and then aggregates the results in 3D space, achieving accurate 3D part segmentation.

\subsection{SegmentNet}
\cref{fig:normal-based_segmentor} (b) illustrates the detailed architecture of SegmentNet.
SegmentNet comprises two processing branches that interact through self-attention mechanisms: one for the normal map (rendered view) and one for the input image (front-view).
The first branch primarily captures geometric information, such as cloth boundaries, from the normal map.
The second branch extracts semantic cues, such as cloth length and style, from the part segments obtained by applying the off-the-shelf image segmentation method (\textit{i.e.}, Sapiens~\cite{khirodkar2024sapiens}) to the input image.
These branches are processed through $L=10$ Transformer layers, where self-attention layers are applied after each Transformer to aggregate features between the two branches.
Then, the first branch originating from the normal map is processed through the remaining Transformer layers to predict the part segments.
All Transformer weights are initialized from the pre-trained Sapiens network~\cite{khirodkar2024sapiens} and are kept frozen during the training of SegmentNet. 
The training details are provided in the supplementary material.

With this design, SegmentNet effectively incorporates features of two branches, resulting in accurate segmentation results.
The normal map (first branch) provides structural details (\textit{e.g.}, human part boundaries) on the rendering viewpoint that include invisible regions from the input image.
However, the normal map can be blurry or noisy due to geometric errors in the textureless human surface.
Thus, relying solely on the normal map can provide wrong segmentations.
To improve the robustness of segmentation, SegmentNet integrates features from visible regions, extracted from the front-view part segments (second branch), which provide semantic cues about human appearance (\textit{e.g.}, cloth style). 
By incorporating the semantic information from the visible region with geometric information from the invisible region, SegmentNet ensures consistent part segmentation even with an imperfect normal map. 
As a result, SegmentNet predicts 3D-consistent and reliable human part labels, enabling \textcolor[RGB]{70,110,255}{\textbf{PartSegmenter}} to produce an accurate 3D part-segmented human surface.

\section{PartTexturer}
\label{sec:PartTexturer}
\textcolor[RGB]{50,190,50}{\textbf{PartTexturer}} reconstructs human textures on the textureless human surface $\textbf{M}_{\text{geo}}$, producing the fully textured 3D human surface $\textbf{M}_{\text{tex}}$.
Based on the part-segmented human surface $\textbf{M}_{\text{part}}$, we distinctly guide each human part's texture using our proposed diffusion model, PartDiffusion.

\subsection{Text captioning}
Before texturing, our framework acquires text prompts that provide descriptive information about the input image.
Each prompt describes attributes, such as hairstyle and cloth color, of $n=5$ human parts.
These text prompts serve as guidance signals for the semantic information of human appearance during the reconstruction of part-specific textures.
In our framework, we use an off-the-shelf text captioning method, BLIP~\cite{li2022blip}.

\subsection{3D human texturing}
To reconstruct the human texture on the textureless human surface $\textbf{M}_{\text{geo}}$, we optimize an MLP network $\phi$ that predicts RGB colors from 3D points on the human surface.
The overall loss function of the optimization is as follows: 
\begin{equation}
    \mathcal{L} = \mathcal{L}_\text{recon} + \mathcal{L}_\text{SDS},
\end{equation}
where $\mathcal{L}_\text{recon}$ is the reconstruction loss, which is computed as the L2 distance between the input image and the rendered image of the human surface from the front-view.
$\mathcal{L}_\text{SDS}$ is a score distillation sampling (SDS) loss function~\cite{poole2022dreamfusion}, which leverages a pre-trained diffusion network to infer invisible human textures from non-frontal views.
Specifically, at a randomly selected camera viewpoint $\textbf{k}$, we render an image $\textbf{x}^{\textbf{k}}$ from human surface colors predicted by the MLP network $\phi$.
Based on the rendered image $\textbf{x}^{\textbf{k}}$, the SDS loss is calculated as
\begin{equation}
\label{eq:sds_loss}
\begin{split}
\nabla_{\phi} \mathcal{L}_{\text{SDS}} = \mathbb{E}[w_t (\hat{\epsilon}_{t}(\mathbf{x}^{\textbf{k}}_{t}; \mathbf{S}^{\textbf{k}}, \textbf{I}, \{P_i\}^{n}_{i=1}) - \epsilon_{t}) \frac{\partial \mathbf{x}^{\textbf{k}}_{t}}{\partial \phi}],
\end{split}
\end{equation} 
where $t$ denotes the noise level, $\mathbf{x}^{\textbf{k}}_{t}$ represents the rendered human surface perturbed by the noise $\epsilon_{t}$, and $w_{t}$ is a weighting factor dependent on the noise level $t$.
$\textbf{S}^{\textbf{k}}$ denotes the part segments obtained by rendering the part-segmented human surface $\textbf{M}_{\text{part}}$, $\textbf{I}$ is the input image, and $\{P_i\}^{n}_{i=1}$ are text prompts obtained from text captioning.
This loss function minimizes the distance between the predicted noise $\hat{\epsilon}_{t}(\cdot)$ and its counterpart noise $\epsilon_{t}$.
Minimizing the distance enforces the rendered human surface $\textbf{x}^{\textbf{k}}$ to match the distribution of human appearances learned by the diffusion network.
In this loss function, a key requirement is for the diffusion network to predict accurate noise while maintaining alignment with the conditioning inputs: $\mathbf{S}^{\textbf{k}}$, $\textbf{I}$, and $\{P_i\}^{n}_{i=1}$.
To this end, we propose PartDiffusion, whose detailed explanation is provided in the following section (\cref{sec:PartDiffusion}).

\subsection{PartDiffusion}
\label{sec:PartDiffusion}
\begin{figure}[t!]
  \centering
  \includegraphics[width=1.0\linewidth]{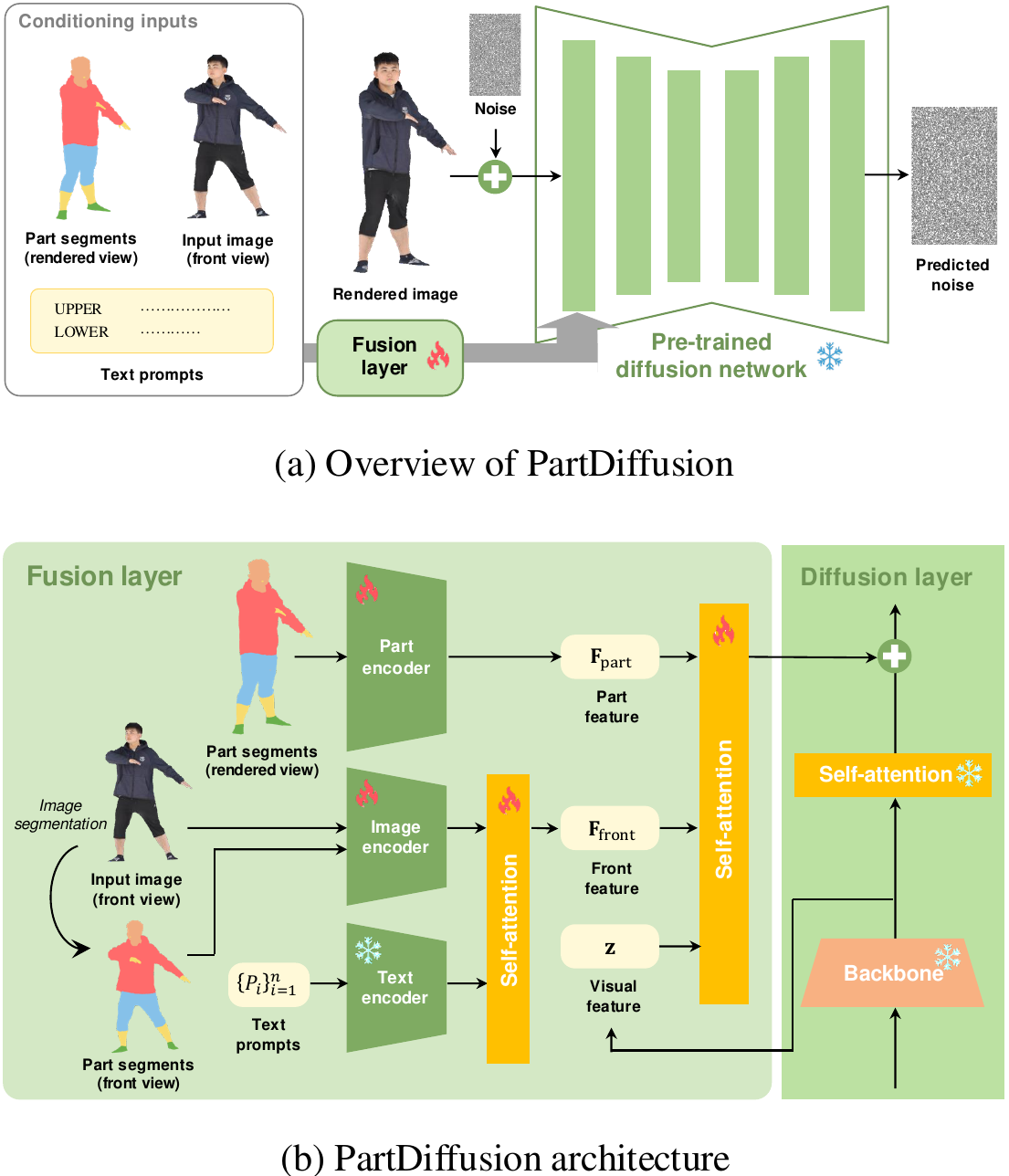}
  \vspace{-1.2em}
  \caption{\textbf{Detailed process of~\textcolor[RGB]{50,190,50}{\textbf{PartTexturer}}.}
    (a) To infer texture for a specific rendering view, PartDiffusion integrates three key sources of information: the front-view image, text prompts, and part segments of the rendering view.
    (b) For effective integration, we adopt self-attention on the separately encoded features.
  }
  \vspace{-0.4em}
  \label{fig:part-aware_diffusion}
\end{figure}

\cref{fig:part-aware_diffusion} (a) illustrates an overview of PartDiffusion.
PartDiffusion takes three conditioning inputs, part segments $\textbf{S}^{\textbf{k}}$, input image $\textbf{I}$, and text prompts $\{P_i\}^{n}_{i=1}$ and predicts the noise corresponding to the noisy image $\textbf{x}^{\textbf{k}}_{t}$, where $\textbf{k}$ is a rendering viewpoint.
To integrate these conditions into the visual features of a pre-trained diffusion network, we aggregate their features via a fusion layer.
All weights of the diffusion network are initialized from StableDiffusion~\cite{rombach2022high} and remain frozen while training PartDiffusion. 
The training details are provided in the supplementary material.

\cref{fig:part-aware_diffusion} (b) shows the detailed process of the fusion layer, which aggregates the three conditioning inputs: part segments (rendered view), input image (front view), and text prompts.
The part segments are encoded into a part feature $\textbf{F}_{\text{part}}$ that captures the human part structural information at the rendered viewpoint.
The input image is processed through the off-the-shelf image segmentation method, Sapiens~\cite{khirodkar2024sapiens}, to extract its part segments (front-view).
Each part region in the part segments is element-wise multiplied with the input image, creating $n=5$ image patches, each containing the appearance of its corresponding part.
The image patches and text prompts are encoded and incorporated separately through a self-attention layer to produce a front feature $\textbf{F}_{\text{front}}$.
This front feature captures the semantic information of each human part in the input image, such as the color and style of the cloth.
The extracted part feature $\textbf{F}_{\text{part}}$ and front feature $\textbf{F}_{\text{front}}$ are aggregated with a visual feature $\textbf{z}$ from the diffusion layer, through a self-attention layer.
The aggregated features are then added to the output of the diffusion layer.
With this architectural design, PartDiffusion effectively integrates the structural information from the part segments in the rendered view and the semantic information from the front-view.
As a result, our proposed PartDiffusion accurately infers human textures that are structurally aligned with human parts and visually consistent with the input image, thereby enabling \textcolor[RGB]{50,190,50}{\textbf{PartTexturer}} to produce high-quality textured human surface $\textbf{M}_{\text{tex}}$.

\section{Experiments}
\subsection{Datasets}
\noindent\textbf{THuman2.1.}
THuman2.1~\cite{tao2021function4d} consists of high-quality 3D scans of 2,445 humans. 
To obtain ground truth (GT) part segmentations for the scans, we apply Sapiens~\cite{khirodkar2024sapiens} to 360 rendered images from uniformly distributed viewpoints and aggregate the results in 3D space.
For our experiments, we uniformly sample 80 scans for evaluation, while the remaining scans are used as training data to train SegmentNet in PartSegmenter and PartDiffusion in PartTexturer.

\noindent\textbf{HuMMan.}
HuMMan~\cite{cai2022humman} is a large-scale dataset containing diverse human poses with 3D scans. 
To obtain GT part segmentations, we apply Sapiens~\cite{khirodkar2024sapiens} to 360 rendered images from uniformly distributed viewpoints and aggregate the results in 3D space.
For our experiments, we uniformly sample 128 scans from the official test set.
This dataset is used only for evaluation purposes.

\noindent\textbf{SHHQ.}
SHHQ~\cite{fu2022stylegan} contains a large number of in-the-wild images showcasing diverse cloth styles and human appearances.
Since SHHQ does not have GT 3D human data, it is used only for qualitative evaluation.

\subsection{Evaluation Metrics.}
\noindent\textbf{3D geometry reconstruction.}
To evaluate the geometry of reconstructed 3D human surfaces, we measure P2S (Point-to-Surface) and CD (Chamfer distance), between the reconstructed and GT human surfaces.
These metrics are measured in centimeters.

\noindent\textbf{3D texture reconstruction.} 
To evaluate the texture of reconstructed 3D human surfaces, we employ PSNR (Peak Signal-to-Noise Ratio), LPIPS~\cite{zhang2018unreasonable}, and Part IoU as evaluation metrics.
To compute these metrics, we first render both the reconstructed and GT human surfaces from pre-defined viewpoints at {0$^{\circ}$, 90$^{\circ}$, 180$^{\circ}$, and 270$^{\circ}$}, where 0$^{\circ}$ represents the front-view.
We then compute PSNR, LPIPS, and Part IoU, between the rendered images.
PSNR and LPIPS quantify the visual fidelity and perceptual similarity of the reconstructed human textures.
Part IoU evaluates the part alignment of the reconstructed human textures in 2D space.
It is computed as the average IoU of part segments of the rendered images, where each segment is obtained by applying Sapiens~\cite{khirodkar2024sapiens} to the rendered images.

\noindent\textbf{3D part segmentation.}
To evaluate 3D part segmentation, we introduce two metrics: Part CD and Label Acc$_{\text{GT}}$.
Part CD evaluates the part alignment of the reconstructed textures in 3D space.
It is computed as the average Chamfer distance across all 3D human parts between the predicted part-segmented human surfaces and their GT counterparts.
We measure Part CD in centimeters.
Label Acc$_{\text{GT}}$ evaluates the accuracy of part segmentation on GT textureless human surface, by eliminating the effect of geometric error.
To compute this, we run PartSegmenter on GT textureless human surfaces and compute the proportion of correctly classified part labels on the surface.
Note that Label Acc cannot be calculated based on reconstructed surfaces, as the reconstructed surfaces have different topologies from the GTs.

\noindent\textbf{2D image generation.}
To verify the learned image priors of the diffusion network used in PartTexturer, we evaluate generated images by the diffusion network, using PSNR and Part IoU.
For evaluation, we obtain 2D part segments by rendering GT human surfaces at pre-defined viewpoints {0$^{\circ}$, 90$^{\circ}$, 180$^{\circ}$, and 270$^{\circ}$}, where 0$^{\circ}$ represents the front-view.
Subsequently, we generate 10 human images corresponding to each 2D part segment.
We compute PSNR and Part IoU between the generated images and their GT counterparts, which are the rendered images from the GT human surfaces.
PSNR evaluates whether the generated images preserve the front-view image's appearance across various viewpoints.
Part IoU evaluates whether the generated images align with the conditioned 2D part segments.

\subsection{Ablation study}
We carry out all ablation studies on THuman2.1~\cite{tao2021function4d} evaluation set.
The geometry reconstruction results are the same across all ablation studies, with P2S and CD being 2.984 and 3.008, respectively.

\begin{figure}[t!]
  \centering
  \includegraphics[width=1.0\linewidth]{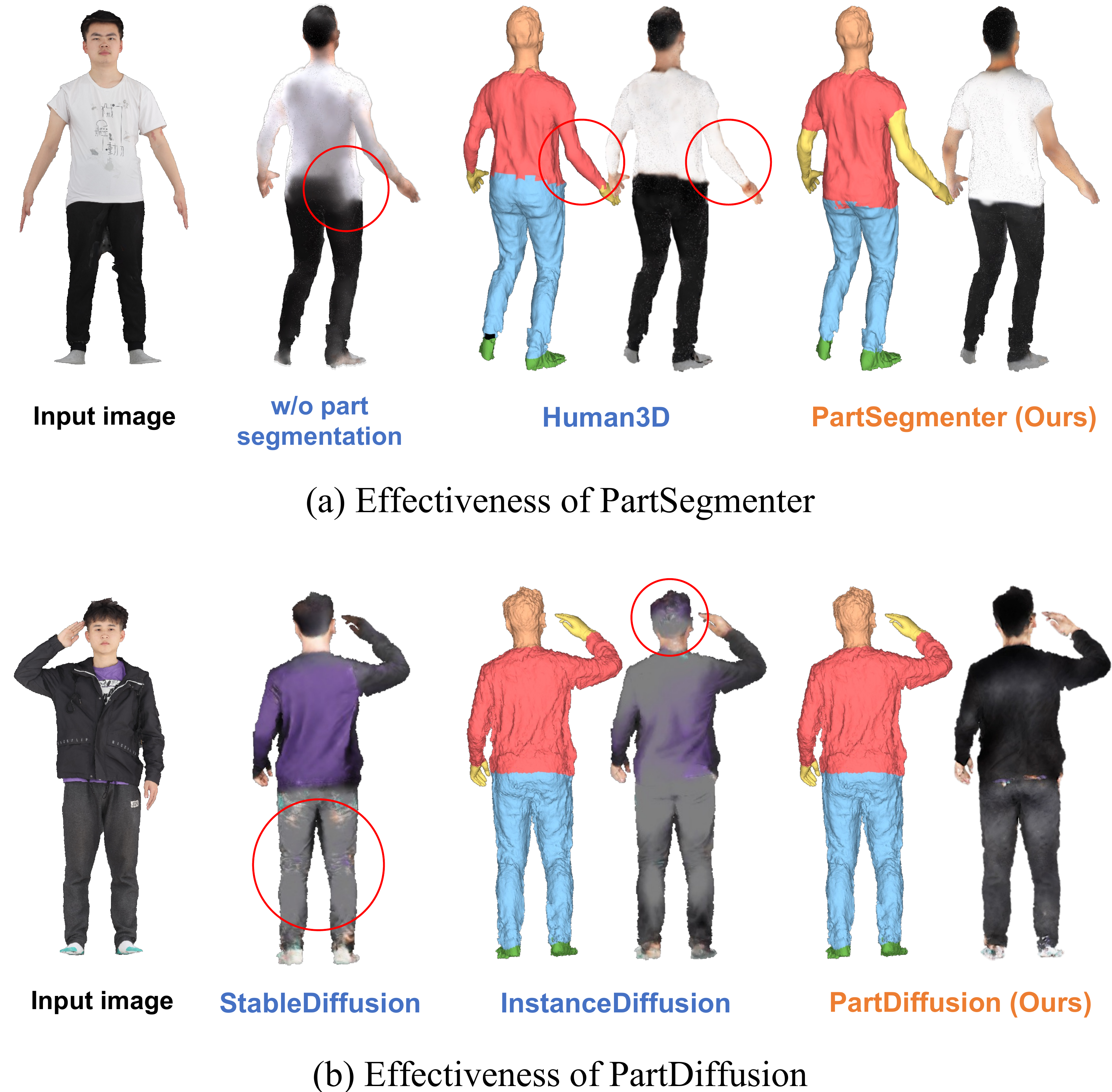}
  \vspace{-1.2em}
  \caption{\textbf{Impact of (a) PartSegmenter and (b) PartDiffusion in 3D human texture reconstruction.}
  }
  \vspace{-0.7em}
  \label{fig:ablation_study}
\end{figure}

\noindent\textbf{Effectiveness of PartSegmenter.}
\cref{fig:ablation_study} (a) and \cref{tab:abla_segmenter} demonstrate the effectiveness of PartSegmenter in both 3D part segmentation and texture reconstruction.
As shown in the first block of \cref{tab:abla_segmenter}, part segmentation information is highly beneficial for texture reconstruction.
Without part segmentation, where the entire human silhouette is treated as a single part segment, the PartSegmenter fails to distinguish different human parts, leading to texture misalignment.
On the other hand, our framework leverages part segmentation as core guidance for texture reconstruction, resulting in accurate textures that are correctly aligned with their corresponding human parts.
Furthermore, our PartSegmenter achieves more accurate 3D part segmentation compared to the existing 3D human segmentation method, Human3D~\cite{takmaz23iccv}.
For a fair comparison, we fine-tuned Human3D on the same training dataset used for PartSegmenter.
Human3D takes a human point cloud as input without utilizing image information.
However, using only geometric information results in imprecise segmentation, as it introduces ambiguity in capturing semantic cues, such as cloth style.
On the other hand, our PartSegmenter utilizes front-view image information along with geometric information (\textit{i.e.}, normal map), effectively capturing semantic cues from the image to support accurate segmentation.
Thus, PartSegmenter achieves superior performance in part segmentation, which contributes to accurate human texture reconstruction.
The second block of \cref{tab:abla_segmenter} shows that incorporating part segments from the front-view image leads to more accurate part segmentation by SegmentNet.
The normal map, which serves as the input to SegmentNet, is often noisy.
To compensate for missing or inaccurate information in the noisy input, our SegmentNet utilizes part segments from the front-view image, which provide additional structural cues.
Accordingly, SegmentNet achieves high part segmentation accuracy and leads to accurate texture reconstruction.

\noindent\textbf{Effectiveness of PartTexturer.}
\cref{fig:ablation_study} (b) and \cref{tab:abla_texturer} demonstrate the effectiveness of PartTexturer in texture reconstruction.
As shown in the first block of \cref{tab:abla_texturer}, our proposed PartDiffusion in PartTexturer is much more effective in both 2D image generation and 3D human texturing than other diffusion networks.
StableDiffusion~\cite{rombach2022high} does not use part segmentation and relies solely on text prompts as its conditioning input.
Without the guidance of part segmentation, it fails to infer the correct textures corresponding to human parts, leading to misaligned texture reconstruction.
InstanceDiffusion~\cite{wang2024instancediffusion}, while taking human segments as input, does not incorporate the information from the front-view image.
Accordingly, it often reconstructs human textures inconsistent with the front-view image's appearance.
On the other hand, our proposed PartDiffusion explicitly conditions the texturing process with both part segmentation and front-view image, ensuring alignment across the human parts in both 2D image generation and 3D texture reconstruction.
The second block of \cref{tab:abla_texturer} shows that utilizing both the front-view image and text prompts leads to visually plausible 2D image generation and accurate texture reconstruction.
The front-view image provides fine-grained human appearance details, ensuring that the reconstructed textures closely match the appearance of the input image.
The text prompts provide semantic information for each part, such as cloth color and style, and contribute to inferring human textures, especially in invisible human regions.
Therefore, by benefiting from PartDiffusion, which effectively aggregates information from the front-view image and text prompts, PartTexturer achieves high-quality and coherent 3D human textures.

\begin{table}[t]
\def\arraystretch{1.45}
\renewcommand{\tabcolsep}{0.8mm}
\footnotesize
\begin{center}
\scalebox{0.77}{
    \begin{tabular}{>{\raggedright\arraybackslash}m{3.7 cm}>{\centering\arraybackslash}m{1.4cm}>{\centering\arraybackslash}m{1.4cm}>{\centering\arraybackslash}m{1.05cm}>{\centering\arraybackslash}m{1.05cm}>{\centering\arraybackslash}m{1.15cm}}
    \specialrule{.1em}{.05em}{0.0em}
        \multicolumn{1}{c|}{} & \multicolumn{2}{c|}{3D part segmentation}  & \multicolumn{3}{c}{3D texture reconstruction} \\
         \multicolumn{1}{l|}{Methods} & Part CD$^{\downarrow}$ & \multicolumn{1}{c|}{Label Acc$^{\uparrow}_{\text{GT}}$} & PSNR$^{\uparrow}$ & LPIPS$^{\downarrow}$ & Part IoU$^{\uparrow}$ \\
        \hline
        \multicolumn{4}{l}{\textbf{$\ast$ Effectiveness of PartSegmenter}} \\
        w/o part segmentation & - & - & 20.992 & 0.115 & 0.461 \\
        Human3D $\dagger$~\cite{takmaz23iccv} & 5.231 & 0.881 & 21.043 & 0.113 & 0.574 \\
        \textbf{PartSegmenter (Ours)} & \textbf{4.369} & \textbf{0.947} & \textbf{22.175} & \textbf{0.096} & \textbf{0.641} \\ \hline
        \multicolumn{4}{l}{\textbf{$\ast$ Variations of SegmentNet design}} \\
        w/o front-view image & 5.087 & 0.941 & 21.654 & 0.105 & 0.635 \\
        \textbf{SegmentNet (Ours)} & \textbf{4.369} & \textbf{0.947} & \textbf{22.175} & \textbf{0.096} & \textbf{0.641} \\
        \specialrule{.1em}{-0.05em}{-0.05em}
    \end{tabular}
}
\end{center}
    \vspace*{-1.2em}
    \caption{
    \textbf{Ablation studies for PartSegmenter.}
    $\dagger$ indicates that the method is fine-tuned on our training datasets for fair comparison.
    }
    \vspace*{-0.5em}
    \label{tab:abla_segmenter}
\end{table}
\begin{table}[t]
\def\arraystretch{1.4}
\renewcommand{\tabcolsep}{0.8mm}
\footnotesize
\begin{center}
\scalebox{0.77}{
    \begin{tabular}{>{\raggedright\arraybackslash}m{3.6cm}>{\centering\arraybackslash}m{1.4cm}>{\centering\arraybackslash}m{1.4cm}>{\centering\arraybackslash}m{1.1cm}>{\centering\arraybackslash}m{1.1cm}>{\centering\arraybackslash}m{1.25cm}}
    \specialrule{.1em}{.05em}{0.0em}
        \multicolumn{1}{c|}{} & \multicolumn{2}{c|}{2D image generation}  & \multicolumn{3}{c}{3D texture reconstruction} \\
         \multicolumn{1}{l|}{Methods} & PSNR$^{\uparrow}$ & \multicolumn{1}{c|}{Part IoU$^{\uparrow}$} & PSNR$^{\uparrow}$ & LPIPS$^{\downarrow}$ & Part IoU$^{\uparrow}$ \\
        \hline
        \multicolumn{4}{l}{\textbf{$\ast$ Effectiveness of PartDiffusion}}  \\ 
        StableDiffusion~\cite{rombach2022high} & 10.719 & 0.138 & 19.276 &  0.166 & 0.472 \\
        InstanceDiffusion~\cite{wang2024instancediffusion} & 16.125 & 0.563 & 20.860 & 0.115 & 0.527 \\
        \textbf{PartDiffusion (Ours)} & \textbf{20.109} & \textbf{0.854} & \textbf{22.175} & \textbf{0.096} & \textbf{0.641} \\ \hline
        \multicolumn{4}{l}{\textbf{$\ast$ Variations of PartDiffusion design}}  \\ 
        w/o front-view image & 19.203 & 0.845 & 21.005 & 0.110 & 0.580 \\
        w/o text prompts & 19.422 & 0.803 & 20.859 & 0.114 & 0.588 \\
        \textbf{PartDiffusion (Ours)} & \textbf{20.109} & \textbf{0.854} & \textbf{22.175} & \textbf{0.096} & \textbf{0.641} \\
        \specialrule{.1em}{-0.05em}{-0.05em}
    \end{tabular}
}
\end{center}
    \vspace*{-1.2em}
    \caption{
    \textbf{Ablation studies for PartTexturer.}
    }
    \vspace*{-0.2em}
    \label{tab:abla_texturer}
\end{table}

\begin{figure*}[t!]
  \centering
  \includegraphics[width=1.0\linewidth]{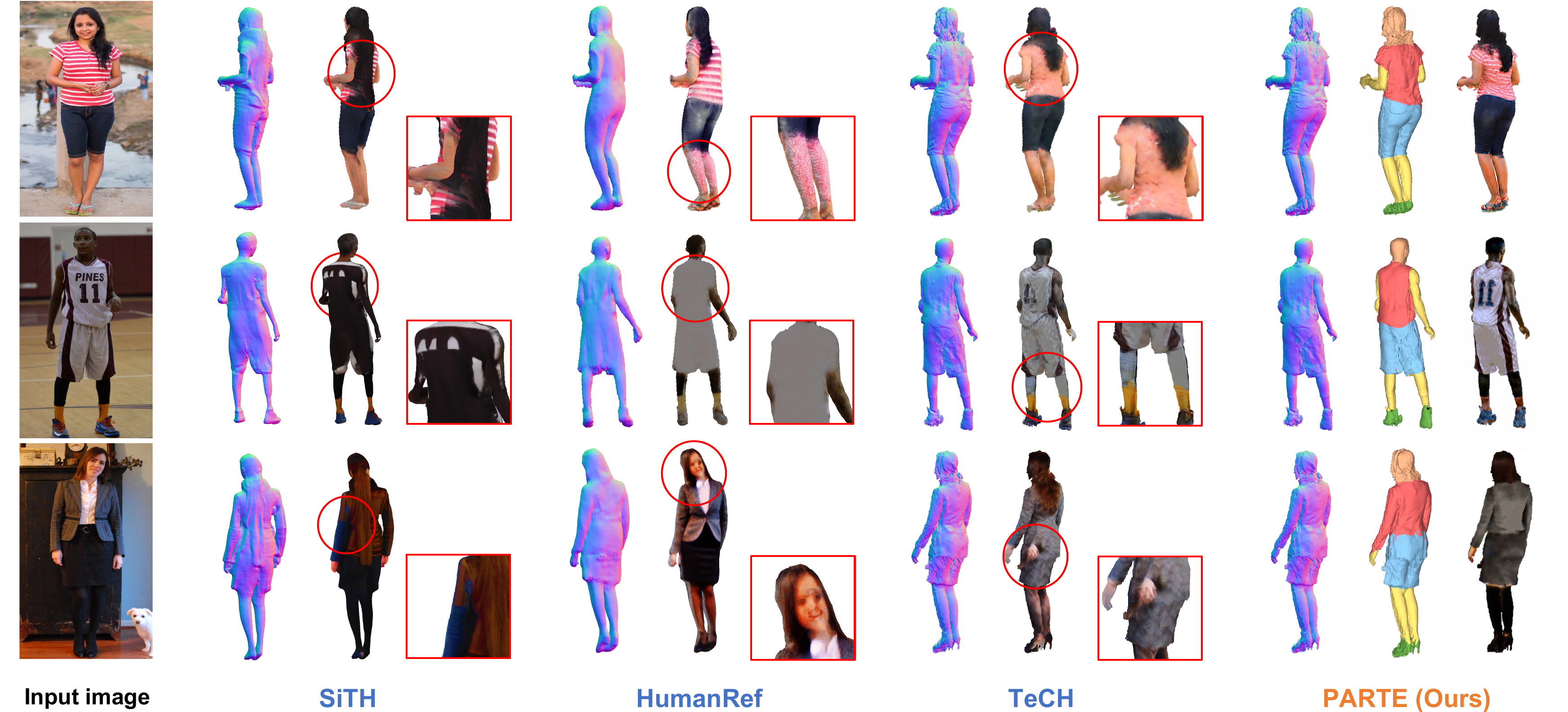}
  \vspace{-1.2em}
\caption{\textbf{Qualitative comparison with existing 3D human reconstruction methods: SiTH~\cite{ho2024sith}, HumanRef~\cite{zhang2024humanref}, and TeCH~\cite{huang2024tech}.}
  }
  \vspace{-0.2em}
  \label{fig:qualitative_comparisons}
\end{figure*}
\begin{table*}[t]
\def\arraystretch{1.28}
\renewcommand{\tabcolsep}{0.8mm}
\small
\begin{center}
\scalebox{0.75}{
    \begin{tabular}{>
    {\raggedright\arraybackslash}m{4.2cm}>{\centering\arraybackslash}m{1.85cm}>{\centering\arraybackslash}m{1.85cm}>{\centering\arraybackslash}m{1.65cm}>{\centering\arraybackslash}m{1.65cm}>{\centering\arraybackslash}m{1.65cm}>{\centering\arraybackslash}m{1.85cm}>{\centering\arraybackslash}m{1.85cm}>{\centering\arraybackslash}m{1.65cm}>{\centering\arraybackslash}m{1.65cm}>{\centering\arraybackslash}m{1.65cm}}
    \specialrule{.1em}{.05em}{0.0em}
     \multicolumn{1}{c|}{}  & \multicolumn{5}{c|}{THuman2.1}  &  \multicolumn{5}{c}{HuMMan} \\ 
         \multicolumn{1}{c|}{} & \multicolumn{2}{c|}{3D geometry reconstruction}  & \multicolumn{3}{c|}{3D texture reconstruction} & \multicolumn{2}{c|}{3D geometry reconstruction}  & \multicolumn{3}{c}{3D texture reconstruction} \\
         \multicolumn{1}{l|}{Methods} & P2S$^{\downarrow}$ & \multicolumn{1}{c|}{CD$^{\downarrow}$} & PSNR$^{\uparrow}$ & LPIPS$^{\downarrow}$ & \multicolumn{1}{c|}{Part IoU$^{\uparrow}$}   & P2S$^{\downarrow}$ & \multicolumn{1}{c|}{CD$^{\downarrow}$} & PSNR$^{\uparrow}$ & LPIPS$^{\downarrow}$ & Part IoU$^{\uparrow}$ \\ \hline 
         \multicolumn{6}{l}{\textbf{$\ast$ Regression-based methods}}  \\ 
        PIFu~\cite{saito2019pifu} &  3.623 & 3.685 & 19.382 & 0.164 & 0.109 & 3.512 & 3.516 & 19.355  & 0.167 & 0.156 \\
        2K2K~\cite{han20232k2k}  & 3.552 & 3.109 & 20.373 & 0.131 & 0.515 & 3.346 & 3.367 & 20.662 & 0.110 & 0.491 \\
        SiTH~\cite{ho2024sith}  & 3.047 & 3.440 & 20.692 & 0.120 & 0.535 & 2.872 & 3.641 & 20.396 & 0.124 & 0.457 \\
        \multicolumn{6}{l}{\textbf{$\ast$ Optimization-based methods}}  \\ 
        HumanRef~\cite{zhang2024humanref}  & 3.185 & 3.199 & 21.302 & 0.113 & 0.576 & 3.312 & 3.591 & 20.974 & 0.101 & 0.442 \\
        SIFU~\cite{zhang2024sifu}  & 3.042 & 3.021 & 21.491 & 0.108 & 0.471 & 2.808 & 3.027 & 21.578 & 0.098 & 0.437 \\
        TeCH~\cite{huang2024tech}  & \textbf{2.984} & \textbf{3.008} & 21.089 & 0.108 & 0.588 & \textbf{2.718} & \textbf{2.667} & 21.642 & 0.097 & 0.437 \\
            \textbf{\ourmethod (Ours)}  & \textbf{2.984} & \textbf{3.008} & \textbf{22.175} & \textbf{0.096} & \textbf{0.641} & \textbf{2.718} & \textbf{2.667} & \textbf{22.076} & \textbf{0.083} & \textbf{0.589}  \\
        \specialrule{.1em}{-0.05em}{-0.05em}
    \end{tabular}
}
\end{center}
    \vspace*{-1.0em}
    \caption{
    \textbf{Quantitative comparisons with existing 3D human reconstruction methods.}
    }
    \vspace*{-0.4em}
    \label{tab:qual_comparison}
\end{table*}
\subsection{Comparison with state-of-the-art methods}
\cref{fig:qualitative_comparisons} and \cref{tab:qual_comparison} show that our \ourmethod~achieves superior texture reconstruction than both regression-based and optimization-based reconstruction methods.
The regression-based methods train a network to infer 3D textures from a single image but struggle to capture the structure of human parts, as they learn global context without explicit local guidance on 3D human parts.
Accordingly, their reconstructed textures often exhibit part misalignment, causing textures to bleed across human parts.
The optimization-based methods, which leverage diffusion networks for high-quality texture reconstruction, also suffer from part misalignment. 
Their diffusion networks heavily rely on text prompts, which describe the overall appearance of the image without distinguishing between different human parts.
Such reliance on text prompts can lead to unintended texturing, as the information from the text prompt is mapped onto incorrect human parts.
Unlike these methods, our framework explicitly incorporates part segmentation from the input image, ensuring that textures are accurately assigned to their corresponding human parts, enhancing the part alignment. 
Additionally, our framework decomposes the texture reconstruction problem into two specialized tasks: part segmentation and part texturing, resulting in high-quality part segmentation and texturing results.
Without explicit part information, the framework is required to infer both the part structure and the overall human texture, making the reconstruction problem significantly more challenging.
In this regard, our framework divides the complex reconstruction process into two stages, PartSegmenter and PartTexturer, each dedicated to the human part structure and human part texture, respectively.
By specializing both stages, our framework enhances the performance of both part segmentation and texture reconstruction, leading to highly detailed and realistic 3D human reconstructions.

\section{Conclusion}
We present \textbf{\ourmethod}, a framework that reconstructs a high-quality 3D human from a single image by explicitly utilizing 3D human part information for reconstructing human texture.
Our framework effectively mitigates texture misalignment across human parts by leveraging our proposed PartSegmenter for 3D part segmentation and PartTexturer for part-guided texturing.
As a result, \ourmethod~achieves state-of-the-art performance in 3D human reconstruction.

\clearpage

\noindent\textbf{Acknowledgements.}
This work was supported in part by the IITP grants [No.2021-0-01343, Artificial Intelligence Graduate School Program (Seoul National University), No.2021-0-02068, No.2023-0-00156, IITP-2025-RS-2020-II201819, RS-2025-02653113, and RS-2025-25441838]. 
This work was also supported by the Industrial Technology Alchemist Project [No. RS-2024-00432410] and the Technology Innovation Program [RS-2025-02653087], both funded by MOTIE, Korea.

\clearpage
\maketitlesupplementary
\setcounter{page}{1}
\setcounter{section}{0}
\setcounter{table}{0}
\setcounter{figure}{0}
\renewcommand{\thesection}{S\arabic{section}}   
\renewcommand{\thetable}{S\arabic{table}}   
\renewcommand{\thefigure}{S\arabic{figure}}

\noindent
\makebox[\textwidth]{%
  \includegraphics[width=0.9\textwidth]{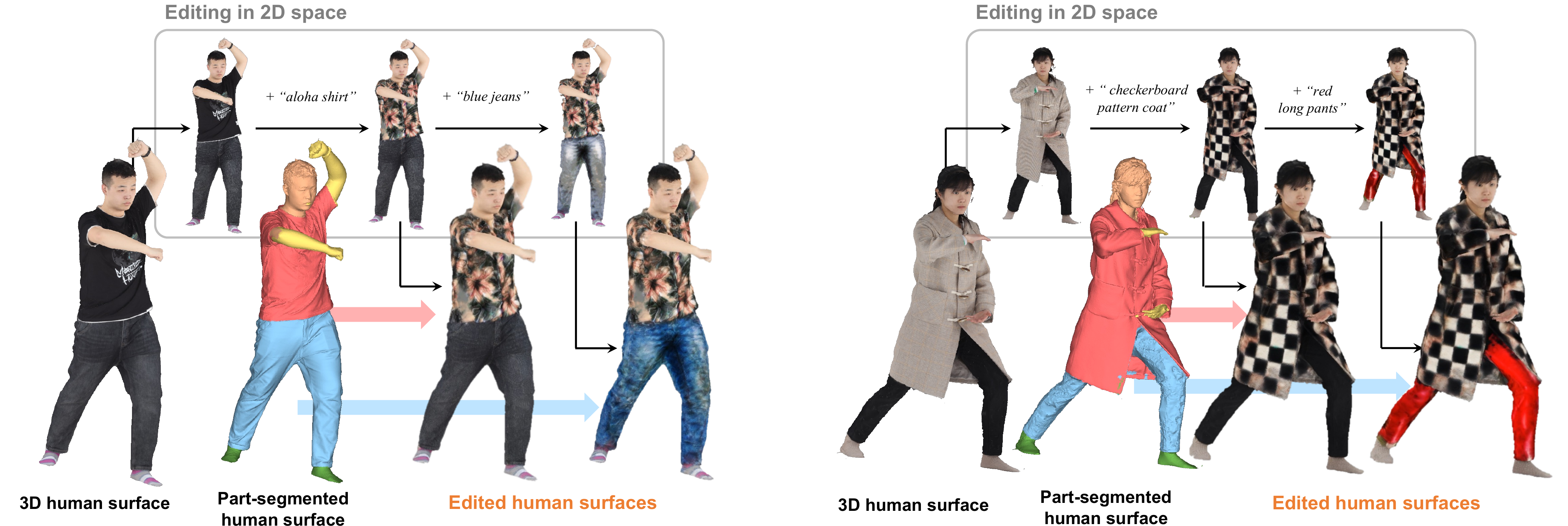}
}
\noindent
\begin{minipage}{\textwidth}
  \captionsetup{justification=centering}
  \vspace{0.8em}
  \captionof{figure}{\textbf{Example of part texture editing.}}
  \label{fig:additional_1}
\end{minipage}

\vspace*{+0.8em}

In this supplementary material, we present additional technical details and more experimental results that could not be included in the main manuscript due to page limitations.
The contents are summarized below:
\begin{itemize}
\vspace{3.0mm}
\setlength{\itemsep}{0pt}
\setlength{\parskip}{0pt}
\setlength{\parsep}{0pt}
\item \ref{sec:additional_application}. Additional applications
\item \ref{sec:plug_in}. Plug-in for other reconstruction methods
\item \ref{sec:gt_mesh}. Texturing based on GT human geometry
\item \ref{sec:more_ablation}. More ablation studies
\item \ref{sec:implementation_details}. Implementation details
\item \ref{sec:more_results}. More comparison results
\item \ref{sec:limitation}. Limitations and future work
\end{itemize}
\vspace{0.8mm}

\section{Additional applications}
\label{sec:additional_application}

\noindent\textbf{Part texture editing.}
\cref{fig:additional_1} shows that our framework allows part-aware texture editing, allowing modifications to a specific human part.
Given a 3D textured human surface and a part-segmented human surface, PartTexturer can edit one of the part textures.
Specifically, after projecting the 3D human surface into 2D space, we can modify the projected image on a target human part via image inpainting methods~\cite{rombach2022high}.
Then, the modified image serves as an input for PartDiffusion, which drives PartTexturer.
By running PartTexturer, we can obtain a new 3D textured human surface where the target part is updated.
Since our proposed PartTexturer takes not only text prompts but also an image as guidance, it enables more precise and detailed 3D editing by referencing the image.

\begin{figure}[t!]
  \centering
  \vspace{+19.0em}
  \includegraphics[width=1.0\linewidth]{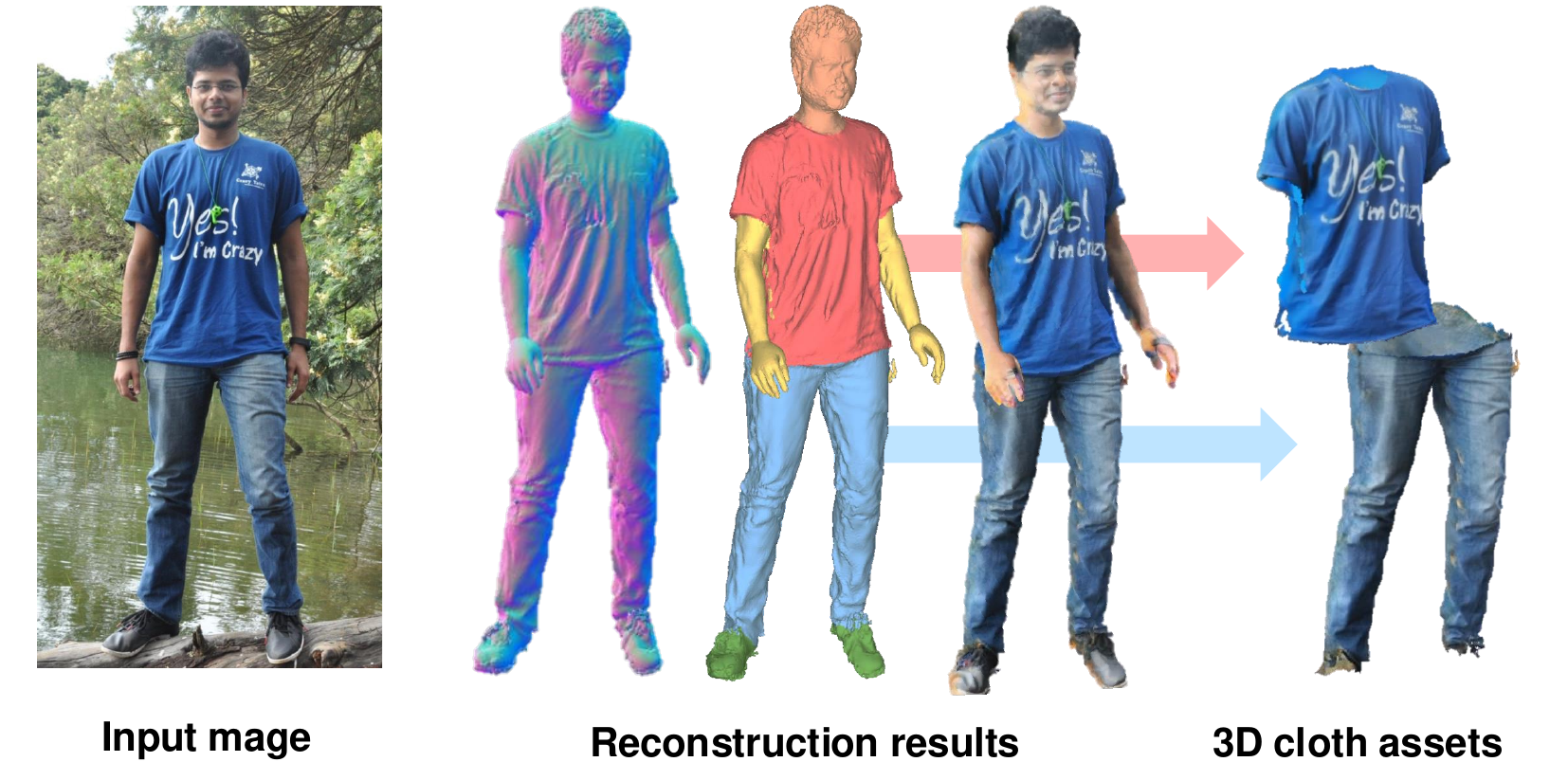}
  \vspace{-1.2em}
  \caption{\textbf{Example of 3D cloth decomposition.}
  }
  \vspace{+0.0em}
  \label{fig:additional_2}
\end{figure}

\noindent\textbf{3D cloth decomposition.}
\cref{fig:additional_2} shows that our framework enables the decomposition of cloth surfaces from the reconstructed result of our framework.
Our framework produces 3D human part segmentation as an intermediate output during reconstruction.
Since reconstructed textures are well-aligned with their corresponding parts, 3D cloth surfaces can be obtained by cutting their regions based on the part segmentation.
This decomposition is made possible because our framework provides accurate human part segmentation and ensures the reconstructed human textures are aligned with the part segmentation.


\section{Plug-in for other reconstruction methods}   
\label{sec:plug_in}
\cref{tab:plugin} shows that integrating our \ourmethod~with various 3D human reconstruction methods, including 2K2K~\cite{han20232k2k}, SiTH~\cite{ho2024sith}, HumanRef~\cite{zhang2024humanref}, and SIFU~\cite{zhang2024sifu}, improves texture quality.
In this experiment, we utilize the geometry outputs from the 3D human reconstruction methods and apply our framework for part-guided texturing. 
In the results, our \ourmethod~achieves superior texture reconstruction compared to the original texturing of each method.
Our framework can be integrated into various 3D human reconstruction pipelines in a plug-and-play manner, enhancing texture quality without altering or modifying the geometry reconstruction process.

\begin{figure}[t!]
  \centering
  \includegraphics[width=1.0\linewidth]{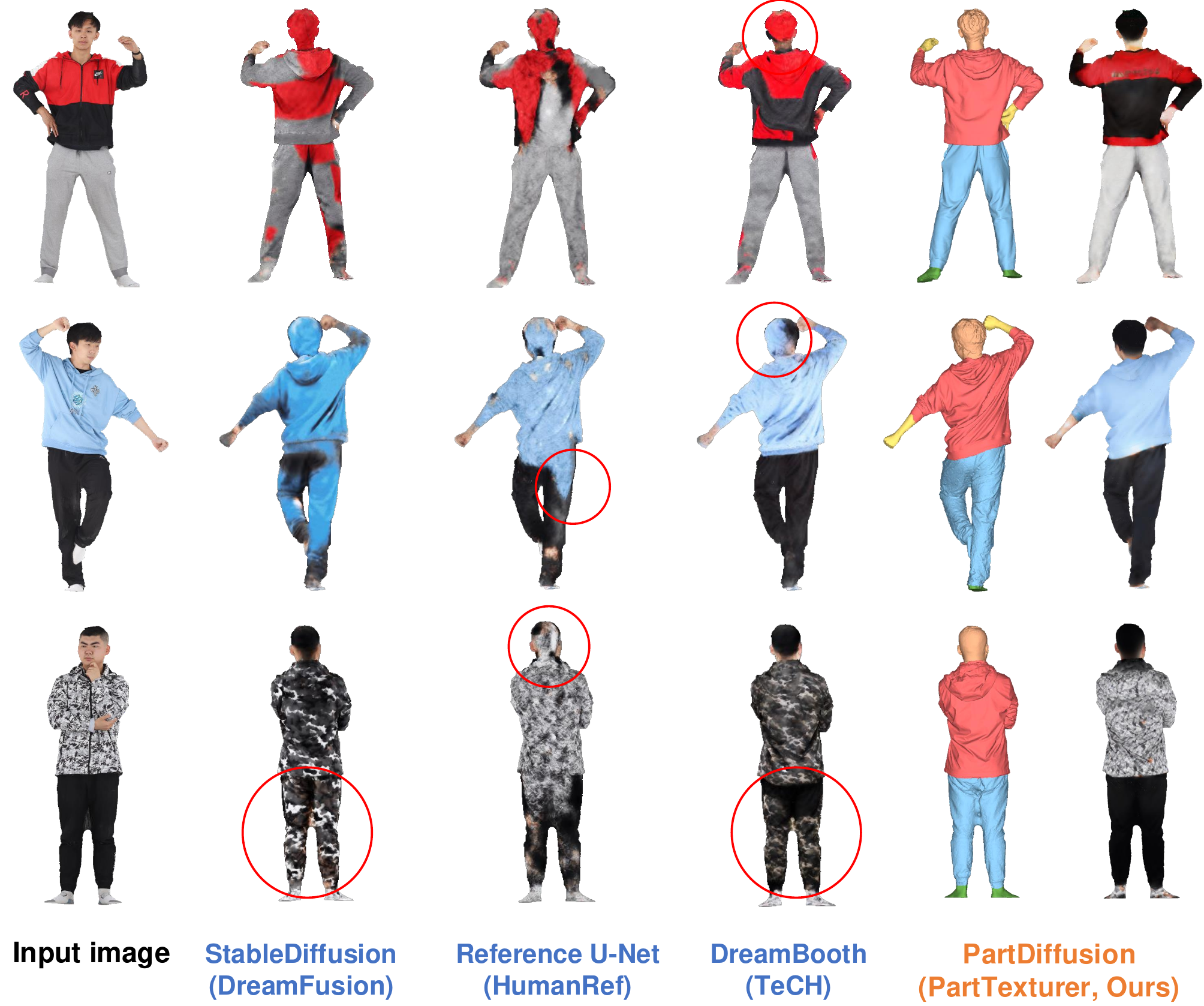}
  \vspace{-1.4em}
  \caption{\textbf{Qualitative comparisons of PartDiffusion with other diffusion models for texturing on GT human geometry.}
  }
  \vspace{-0.0em}
  \label{fig:gt_mesh}
\end{figure}
\section{Texturing based on GT human geometry}
\label{sec:gt_mesh}
\cref{fig:gt_mesh} and \cref{tab:gt_mesh} demonstrate that our framework achieves superior texture reconstruction compared to other diffusion models when evaluated on GT human geometry of THuman2.1~\cite{tao2021function4d}.
To evaluate texture reconstruction while eliminating the influence of geometric errors, we compare our framework with other texturing approaches based on GT human geometries.
Specifically, we remove textures from the GT human geometries of THuman2.1 and use them as inputs to texturing frameworks.
As a result, our framework outperforms other diffusion models in both texture fidelity and alignment across human parts.

\section{More ablation studies}
\label{sec:more_ablation}
\begin{figure*}[t!]
  \centering
  \includegraphics[width=0.88\linewidth]{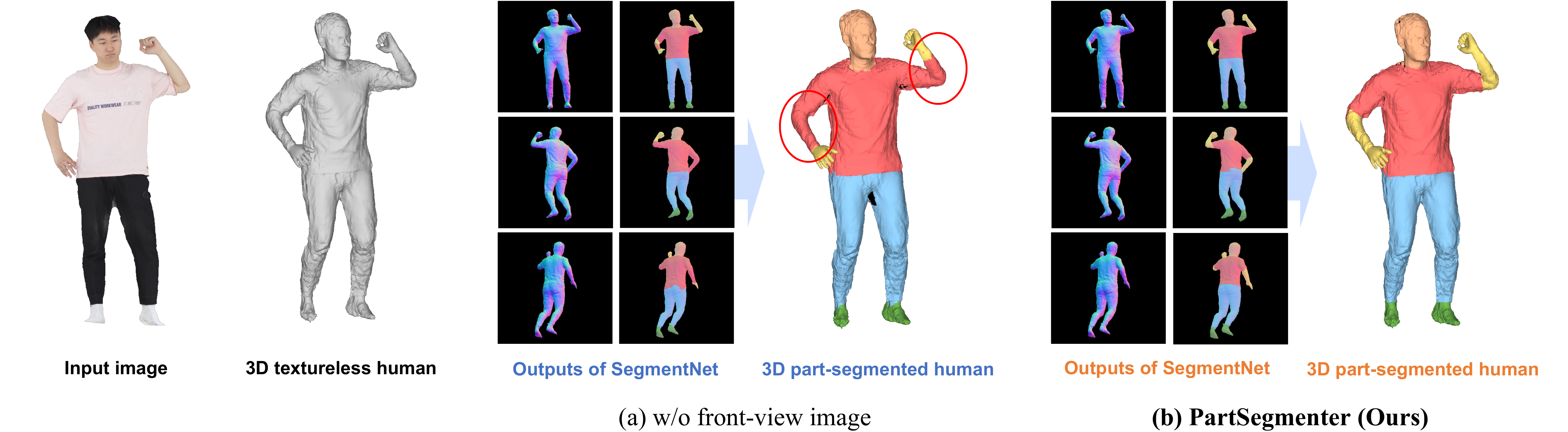}
  \vspace{+0.0em}
  \caption{\textbf{Ablation study for SegmentNet design.}
  }
  \vspace{+0.4em}
  \label{fig:suppl_ablation_segmenter}
\end{figure*}

\noindent\textbf{Effectiveness of SegmentNet design.}
\cref{fig:suppl_ablation_segmenter} shows that incorporating front-view part segments enhances the part segmentation.
The reconstructed 3D textureless human surface exhibits indistinct boundaries between different human part regions, making it challenging to accurately segment each part.
Accordingly, without front-view part segments, SegmentNet produces incorrect 2D part segments, which lead to failures in 3D part segmentation.
To address this, we incorporate front-view part segments for the segmentation, which capture semantic cues that are not explicitly represented in the normal map.
By leveraging these additional semantic cues, our approach enables more accurate part segmentation.

\begin{table}[t]
\def\arraystretch{1.4}
\renewcommand{\tabcolsep}{0.8mm}
\footnotesize
\begin{center}
\scalebox{0.84}{
    \begin{tabular}{>{\raggedright\arraybackslash}m{4.8cm}|>{\centering\arraybackslash}m{1.45cm}>{\centering\arraybackslash}m{1.45cm}>{\centering\arraybackslash}m{1.45cm}}
    \specialrule{.1em}{.05em}{0.0em}
        &  \multicolumn{3}{c}{Texture reconstruction} \\
         \multicolumn{1}{l|}{Methods} & PSNR$^{\uparrow}$ & LPIPS$^{\downarrow}$ & Part IoU$^{\uparrow}$ \\
        \hline
        2K2K~\cite{han20232k2k} & 20.373 & 0.131 & 0.515 \\
        \textbf{2K2K~\cite{han20232k2k} + \ourmethod (Ours)}  & \textbf{20.692} & \textbf{0.128} & \textbf{0.574} \\ \hline
        SiTH~\cite{ho2024sith} & 20.692 & 0.120 & 0.535 \\
        \textbf{SiTH~\cite{ho2024sith} + \ourmethod (Ours)}  & \textbf{21.449} & \textbf{0.108} & \textbf{0.585} \\ \hline
        HumanRef~\cite{zhang2024humanref} & 21.302 & 0.113 & 0.576 \\
        \textbf{HumanRef~\cite{zhang2024humanref} + \ourmethod (Ours)}  & \textbf{22.153} & \textbf{0.101} & \textbf{0.623} \\ \hline
        SIFU~\cite{zhang2024sifu} & 21.491 & 0.108 & 0.588 \\
        \textbf{SIFU~\cite{zhang2024sifu} + \ourmethod (Ours)}  & \textbf{22.412} & \textbf{0.095} & \textbf{0.639} \\ \hline
        TeCH~\cite{huang2024tech} & 21.089 & 0.108 & 0.588 \\
        \textbf{TeCH~\cite{huang2024tech} + \ourmethod (Ours)}  & \textbf{22.175} & \textbf{0.096} & \textbf{0.641}\\
        \specialrule{.1em}{-0.05em}{-0.05em}
    \end{tabular}
}
\end{center}
    \vspace*{-1.0em}
    \caption{
    \textbf{Impact of applying \ourmethod~to different 3D reconstruction methods on THuman2.1~\cite{tao2021function4d}.}
    }
    \vspace*{+0.2em}
    \label{tab:plugin}
\end{table}
\begin{table}[t]
\def\arraystretch{1.4}
\renewcommand{\tabcolsep}{0.8mm}
\footnotesize
\begin{center}
\scalebox{0.84}{
    \begin{tabular}{>{\raggedright\arraybackslash}m{4.8cm}|>{\centering\arraybackslash}m{1.45cm}>{\centering\arraybackslash}m{1.45cm}>{\centering\arraybackslash}m{1.45cm}}
    \specialrule{.1em}{.05em}{0.0em}
        &  \multicolumn{3}{c}{Texture reconstruction} \\
         \multicolumn{1}{l|}{Methods} & PSNR$^{\uparrow}$ & LPIPS$^{\downarrow}$ & Part IoU$^{\uparrow}$ \\
        \hline
        StableDiffusion~\cite{rombach2022high} (DreamFusion~\cite{poole2022dreamfusion}) & 27.422 & 0.048 & 0.772 \\
        Reference U-Net (HumanRef~\cite{zhang2024humanref}) & 27.659 & 0.042 & 0.815 \\
        DreamBooth~\cite{ruiz2023dreambooth} (TeCH~\cite{huang2024tech}) & 28.337 & \textbf{0.039} & 0.835 \\
        \textbf{PartDiffusion (PartTexturer, Ours)} & \textbf{29.315} & \textbf{0.039} & \textbf{0.857} \\ 
        \specialrule{.1em}{-0.05em}{-0.05em}
    \end{tabular}
}
\end{center}
    \vspace*{-1.2em}
    \caption{
    \textbf{Comparisons of texturing results between different diffusion models based on textureless GT human geometry of THuman2.1~\cite{tao2021function4d}.}
    }
    \vspace*{+0.0em}
    \label{tab:gt_mesh}
\end{table}

\begin{figure*}[t!]
  \centering
  \includegraphics[width=1.0\linewidth]{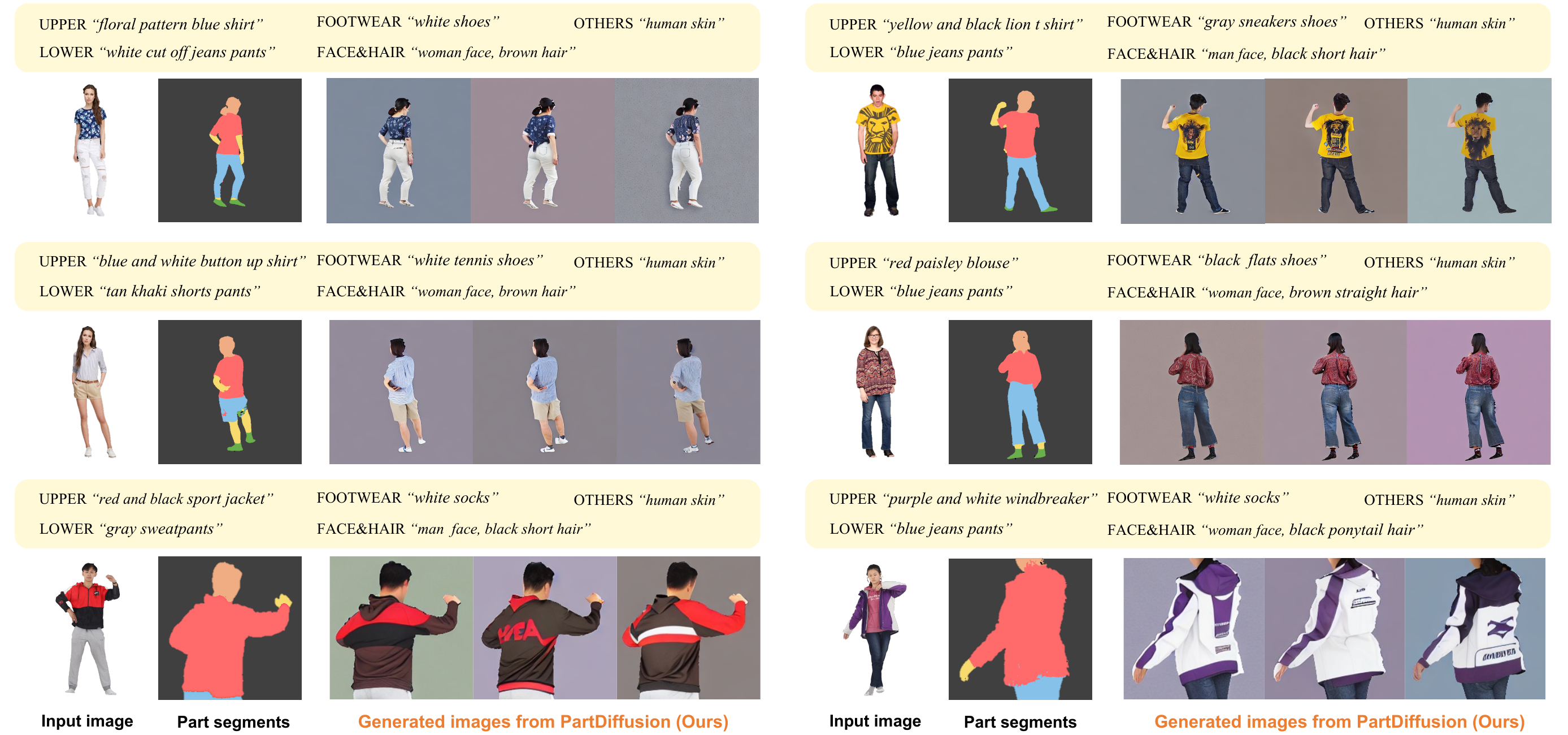}
  \vspace{-1.4em}
  \caption{\textbf{Image generation examples of PartDiffusion.}
  }
  \vspace{+0.0em}
  \label{fig:diffusion_results}
\end{figure*}

\begin{figure*}[!tb]
  \centering
  \includegraphics[width=0.95\linewidth]{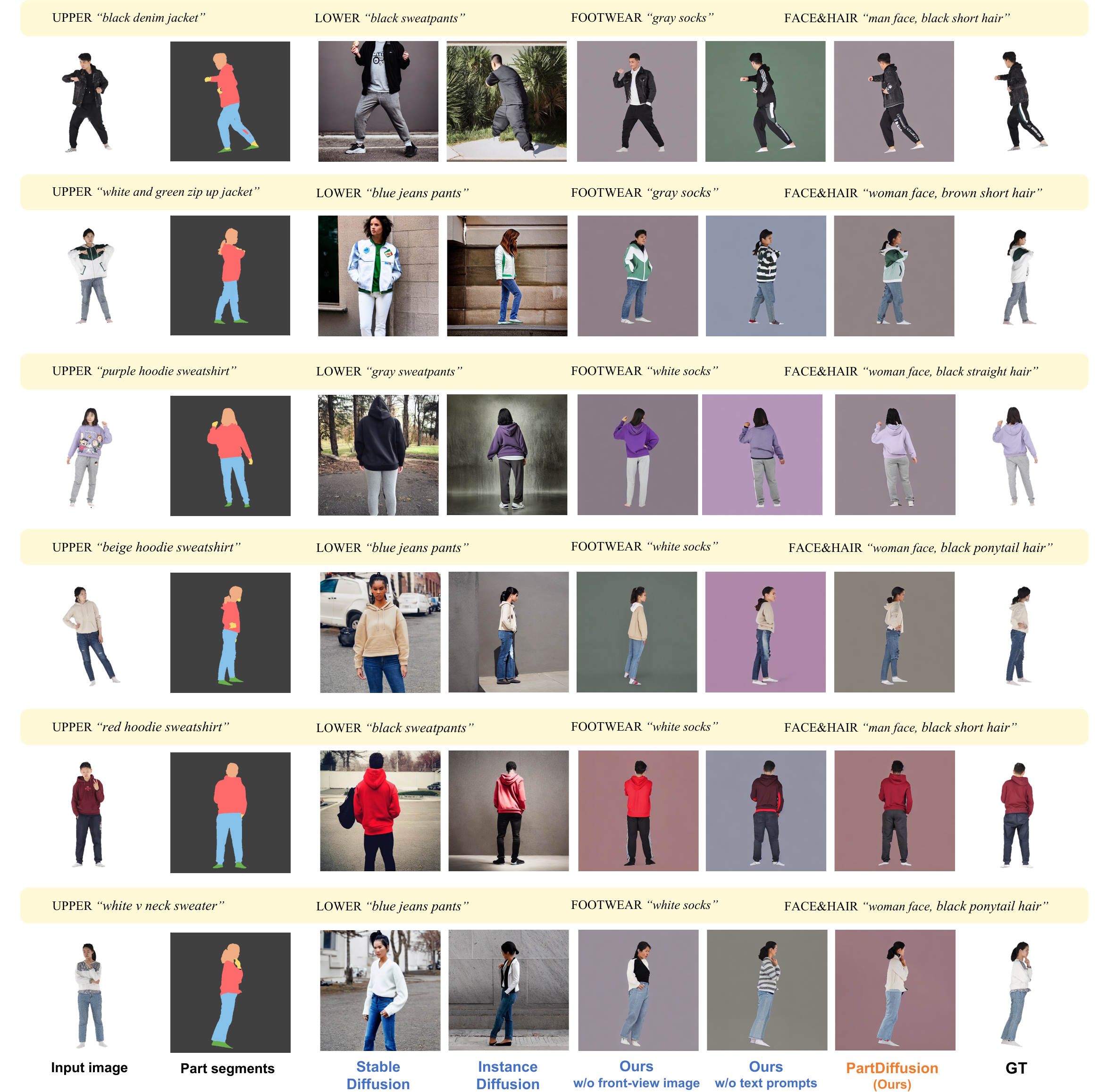}
  \vspace{-0.2em}
  \caption{\textbf{Qualitative comparison of image generation with various diffusion networks and PartDiffusion, on THuman2.1~\cite{tao2021function4d}.}
  }
  \vspace{-0.0em}
  \label{fig:suppl_diffusion_ablation}
\end{figure*}

\noindent\textbf{Effectiveness of PartDiffusion design.}
\cref{tab:image_generation} and \cref{fig:suppl_diffusion_ablation} show that PartDiffusion effectively generates human images that are accurately aligned with both the input image and part segments, compared to other diffusion networks.
For quantitative comparison, we measure PSNR, LPIPS, and Part IoU between generated images and GT counterparts.
All other diffusion networks except PartDiffusion struggle to preserve both the human part structure and human appearance from the input image.
This limitation leads to inconsistent 3D human texturing, resulting in misaligned textures across human parts.
In contrast, our PartDiffusion effectively integrates the input image and part segments, ensuring proper part alignment while generating visually coherent human images.
This indicates that PartDiffusion possesses precise prior knowledge of both human part structure and human appearance, enabling more accurate 3D human texturing in PartTexturer.
\cref{fig:diffusion_results} additionally shows that PartDiffusion is capable of generating human images while preserving the appearance of the input image, even when given in-the-wild images with diverse clothing styles.


\section{Implementation details}
\label{sec:implementation_details}
We provide an explanation of the implementation details of PartSegmenter and PartTexturer below.
PyTorch~\cite{paszke2017automatic} is used for all implementations.

\subsection{PartSegmenter}
\noindent\textbf{Geometry reconstruction.}
To reconstruct a 3D textureless human surface from a single image, we employ an off-the-shelf reconstruction method, TeCH~\cite{huang2024tech}.
TeCH uses Deep Marching Tetrahedra~\cite{shen2021deep} (DMTet) as a geometric representation of a 3D human.
DMTet represents 3D geometry based on a tetrahedral grid structure, where each 3D query point on the tetrahedral grid predicts a signed distance from the 3D geometry surface.
Based on the 3D representation, we initially optimize it based on the naked human body, SMPL-X~\cite{pavlakos2019expressive} human mesh, which is estimated from PIXIE~\cite{feng2021collaborative}.
After initialization, the geometry is optimized to capture fine human details, with three types of losses: reconstruction loss, SDS loss, and regularization loss.
Reconstruction loss is defined as the L2 distance between the normal rendering results of DMTet and the predicted normal maps based on Sapiens~\cite{khirodkar2024sapiens} normal estimator.
SDS loss enforces the geometry's normal rendering results to match the real image knowledge learned by the diffusion model~\cite{rombach2022ldm}.
Regularization loss inhibits implausible geometry through Laplacian smoothing~\cite{ando2006learning}.
We used Adam~\cite{kingma2015adam} optimizer with a base learning rate $1 \times 10^{-3}$ with a weight decay of $5 \times 10^{-4}$.
The optimization was done for 10,000 steps with a single NVIDIA A100 40GB GPU.
After the optimization, we convert the DMTet into a textureless 3D human surface with Marching Tetrahedra (MT)~\cite{doi1991efficient} algorithm.

\noindent\textbf{3D part segmentation.}
For 3D part segmentation from a 3D textureless human surface, we first render multiple normal maps from 30 uniformly distributed viewpoints.
Then, the normal maps are forward into SegmentNet to obtain part segments corresponding to the viewpoints.
For the front viewpoint, which aligns with the input image, we utilize the image segmentation method Sapiens~\cite{khirodkar2024sapiens} instead of SegmentNet.
The pixel labels of the part segments are unprojected onto the 3D human surface and used for voting. 
By aggregating the 30 part segments, we assign each surface vertex the most frequently occurring part label as the final label, resulting in a 3D part-segmented human surface.

\noindent\textbf{Training details of SegmentNet.}
Our SegmentNet is designed by modifying the off-the-shelf image segmentation network, Sapiens-1b~\cite{khirodkar2024sapiens}.
We apply the publicly released pre-trained weights to all Transformer layers of SegmentNet while keeping them frozen.
Then, we insert self-attention layers after the first $L=10$ Transformer layers out of the total $40$ layers in Sapiens.
For training SegmentNet, we utilize weighted cross-entropy loss by following Sapiens.
Data augmentation, including scaling, rotation, flipping, and color jittering, is performed in training.
The weights are updated by AdamW~\cite{loshchilov2019adamw} optimizer with a batch size of 2. 
The initial learning rate is set to $5 \times 10^{-4}$
and linearly reduced to $0$ over training.
We train SegmentNet for 5 epochs with a single NVIDIA A100 40GB GPU.

\subsection{PartTexturer}
\noindent\textbf{3D human texturing.}
In 3D human texturing, we optimize a MLP network that predicts a RGB color value at the input 3D coordinate.
The MLP network is implemented by using a fully-connected layer with 32 hidden dimension and ReLU activations.
It takes 3D coordinates of the human surface as input, after applying the hash positional encoding with a maximum resolution of 2048.
In \cref{eq:sds_loss}, we use the classifier-free guidance~\cite{ho2021cfg} strategy with a guidance scale of 100 for noise estimation.
The noise levels are defined at randomly selected timesteps within the range $[0.02, 0.98]$.
We used Adam~\cite{kingma2015adam} optimizer to optimize the network with an exponentially decaying learning rate starting from $1 \times 10^{-2}$.
We optimize the network for 4,000 steps with a batch size 4 on a single NVIDIA A100 40GB GPU.

\begin{table}[t]
\def\arraystretch{1.4}
\renewcommand{\tabcolsep}{0.8mm}
\footnotesize
\begin{center}
\scalebox{0.78}{
    \begin{tabular}{>{\raggedright\arraybackslash}m{5.5 cm}|>{\centering\arraybackslash}m{1.45cm}>{\centering\arraybackslash}m{1.45cm}>{\centering\arraybackslash}m{1.45cm}}
    \specialrule{.1em}{.05em}{0.0em}
        &  \multicolumn{3}{c}{Image generation} \\
         \multicolumn{1}{l|}{Methods} & PSNR$^{\uparrow}$ &  LPIPS$^{\downarrow}$ & Part IoU$^{\uparrow}$ \\
        \hline
        StableDiffusion~\cite{rombach2022high} & 10.719 & 0.491 & 0.138 \\
        InstanceDiffusion~\cite{wang2024instancediffusion} & 16.125 & 0.202 &  0.563\\
        \textbf{PartDiffusion} w/o image segments & 15.391 & 0.234 & 0.492 \\
        \textbf{PartDiffusion} w/o text prompts & 19.422 & 0.134 & 0.803 \\
        \textbf{PartDiffusion (Ours)} & \textbf{20.109} & \textbf{0.119} & \textbf{0.854} \\ 
        \specialrule{.1em}{-0.05em}{-0.05em}
    \end{tabular}
}
\end{center}
\vspace*{-1.0em}
\caption{
\textbf{Ablation studies on image generation quality among different diffusion networks on THuman2.1~\cite{tao2021function4d}.}
}
\vspace*{-0.6em}
\label{tab:image_generation}
\end{table}
\begin{table}[t]
\def\arraystretch{1.4}
\renewcommand{\tabcolsep}{0.8mm}
\footnotesize
\begin{center}
\scalebox{0.95}{
    \begin{tabular}{>{\raggedright\arraybackslash}m{2.9cm}|>{\centering\arraybackslash}m{1.15cm}>{\centering\arraybackslash}m{1.15cm}>{\centering\arraybackslash}m{1.4cm}}
    \specialrule{.1em}{.05em}{0.0em}
    Methods & PSNR$^{\uparrow}$ & LPIPS$^{\downarrow}$ & Part IoU$^{\uparrow}$ \\
    \hline
    SiTH [\textcolor{iccvblue}{23}] & 20.200 & 0.155 & 0.480 \\
    HumanRef [\textcolor{iccvblue}{80}] & 20.896 & 0.143 & 0.442 \\
    TeCH [\textcolor{iccvblue}{29}] & 21.090 & 0.123 & 0.489 \\
    \textbf{PARTE (Ours)} & \textbf{21.698} & \textbf{0.113} & \textbf{0.512} \\
    \specialrule{.1em}{-0.05em}{-0.05em}
    \end{tabular}   
}
\end{center}
\vspace*{-1.0em}
\caption{
\textbf{Quantitative comparisons with existing 3D human reconstruction methods, on 4D-DRESS~\cite{wang20244d}.}
}
\vspace*{+0.0em}
\label{tab:4d_dress}
\end{table}

\noindent\textbf{Training details of PartDiffusion.}
We use three types of encoders in our PartDiffusion.
Part encoder uses a ConvNeXt-T~\cite{liu2022convnext} architecture following~\cite{wang2024instancediffusion}.
For the image encoder, we design a new module which consists of 4 ConvNeXt blocks, where the layer depths of each block are $[3, 3, 3, 1]$ and the feature sizes of each block being $[16, 32, 64, 12]$.
The prompt encoder follows the structure of CLIP~\cite{radford2021clip} encoder.
For training PartDiffusion, we adopt the pre-trained weights of InstanceDiffusion~\cite{wang2024instancediffusion} as the initial weights for the part encoder and the subsequent self-attention layer.
The prompt encoder and all layers of the diffusion network are initialized with pre-trained CLIP~\cite{radford2021clip} and StableDiffusion~\cite{rombach2022high}, respectively, and are kept frozen during training.
To train the network, we acquire sets of front-view images, novel-view images, novel-view part segments, and text prompts.
The front- and novel-view images are obtained by rendering 3D human scans from two randomly selected viewpoints.
Then, novel-view part segments are extracted from the novel-view images using Sapiens~\cite{khirodkar2024sapiens}.
The text prompts are automatically generated from the front-view images using the off-the-shelf text captioning model BLIP~\cite{li2022blip}.
Using these data, we train the network by minimizing the L2 distance between the estimated noise and the target noise, following the conventional training strategy of diffusion networks.
AdamW~\cite{loshchilov2019adamw} optimizer is used for the training with a base learning rate of $5 \times 10^{-5}$.
We train PartDiffusion for 36,000 steps with a batch size of 4 on a single NVIDIA A100 40GB GPU.

\begin{figure}[t!]
  \centering
  \includegraphics[width=0.68\linewidth]{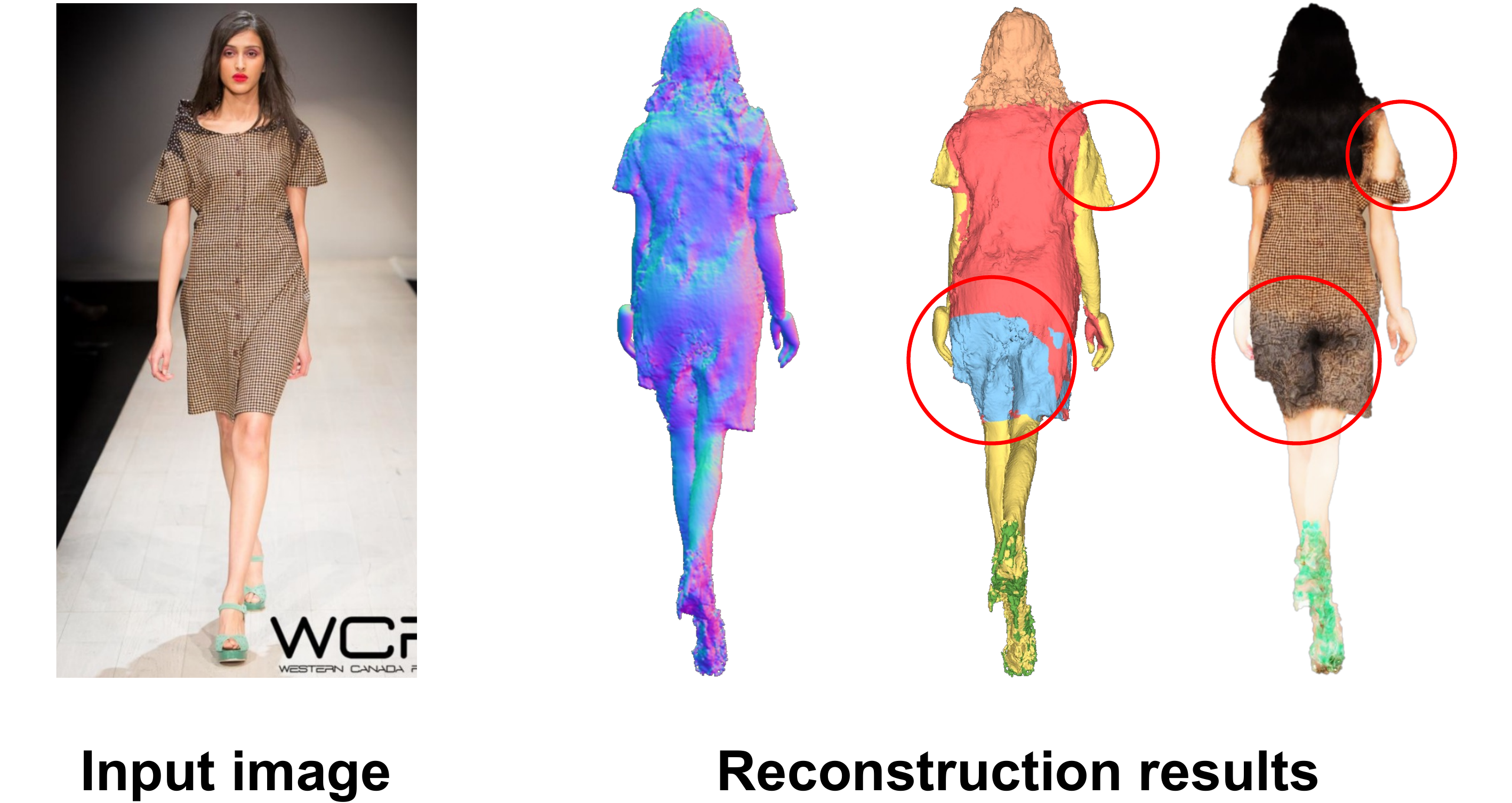}
  \vspace{-0.0em}
  \caption{\textbf{Failure case of our proposed framework.}
  }
  \label{fig:suppl_limitation}
\end{figure}
\section{More comparison results}
\label{sec:more_results}
\begin{figure*}[t!]
  \centering
\includegraphics[width=1.0\linewidth]{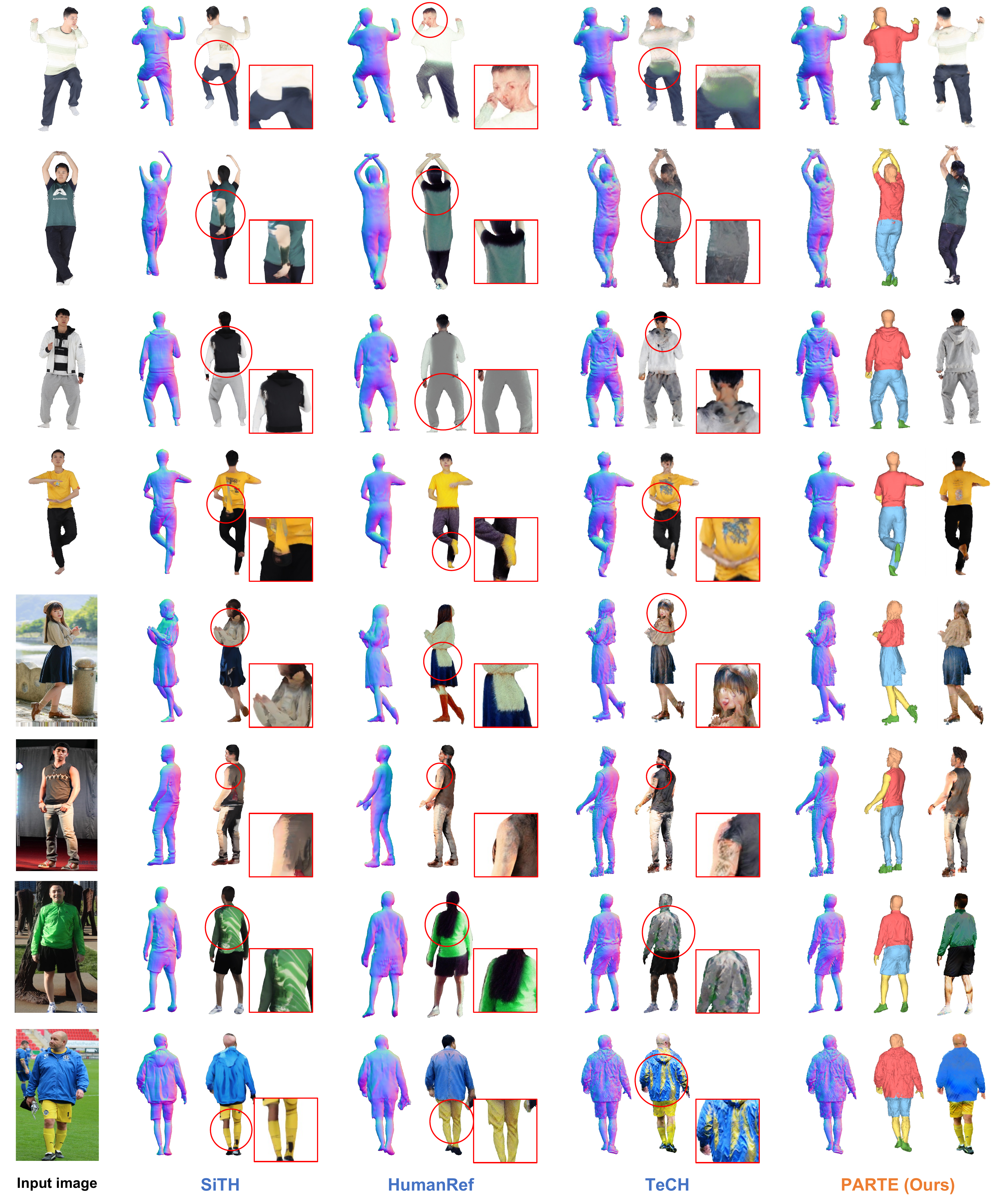}
  \vspace{-1.2em}
  \caption{\textbf{Qualitative comparison of~\ourmethod~ with SiTH~\cite{ho2024sith}, HumanRef~\cite{zhang2024humanref}, and TeCH~\cite{huang2024tech}, on THuman2.1~\cite{tao2021function4d} and SHHQ~\cite{fu2022stylegan}.}
  We highlight their representative failure cases with red circles.
  }
  \vspace{-0.4em}
  \label{fig:more_comparison}
\end{figure*}

\begin{figure*}[t]
  \centering
  \includegraphics[width=0.88\linewidth]{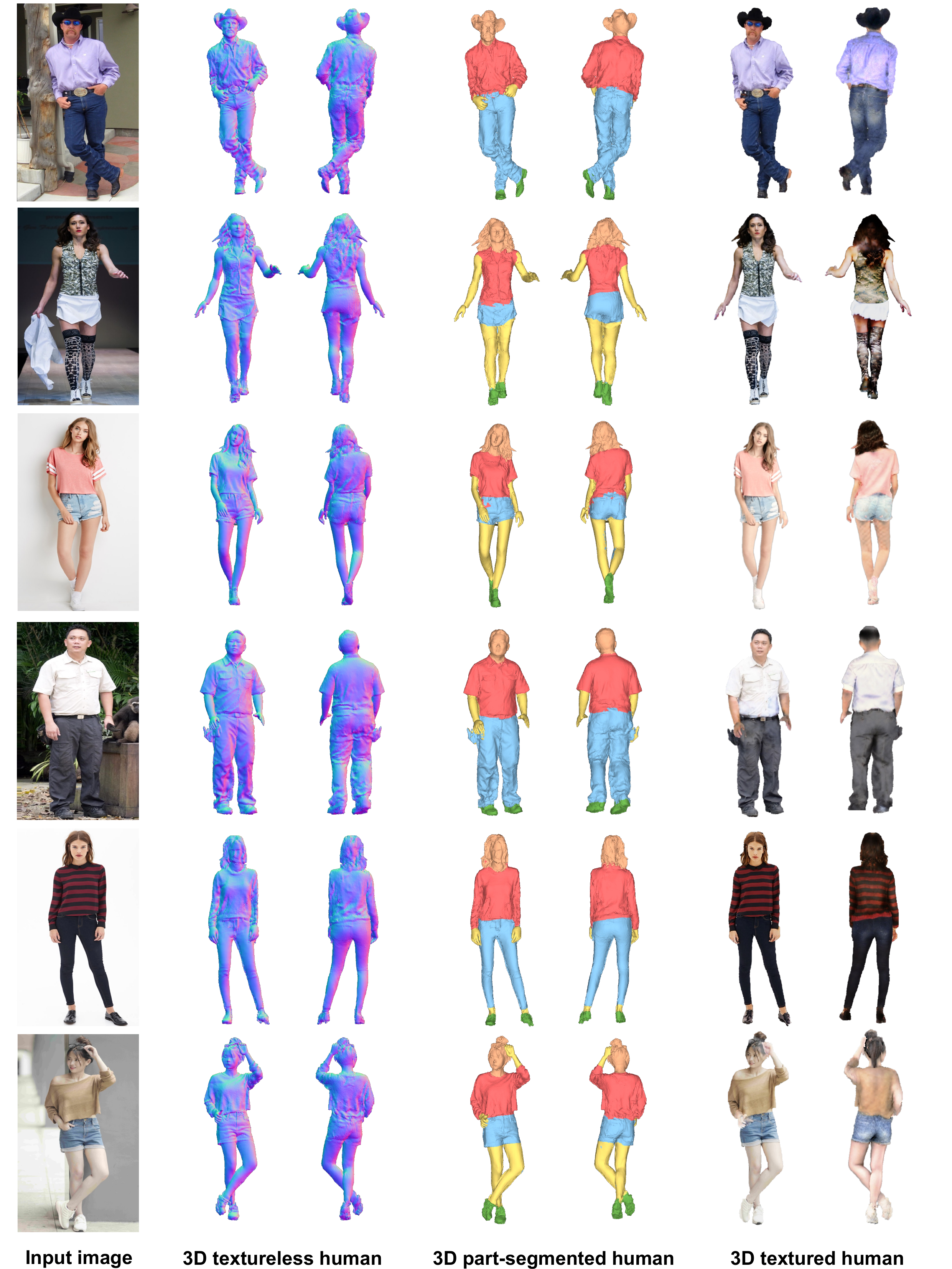}
  \vspace{+0.2em}
  \caption{\textbf{More qualitative results of \ourmethod~on in-the-wild images of SHHQ~\cite{fu2022stylegan}.}
  }
  \vspace{+1.0em}
  \label{fig:more_results_1}
\end{figure*}

\begin{figure*}[t]
  \centering
  \includegraphics[width=0.88\linewidth]{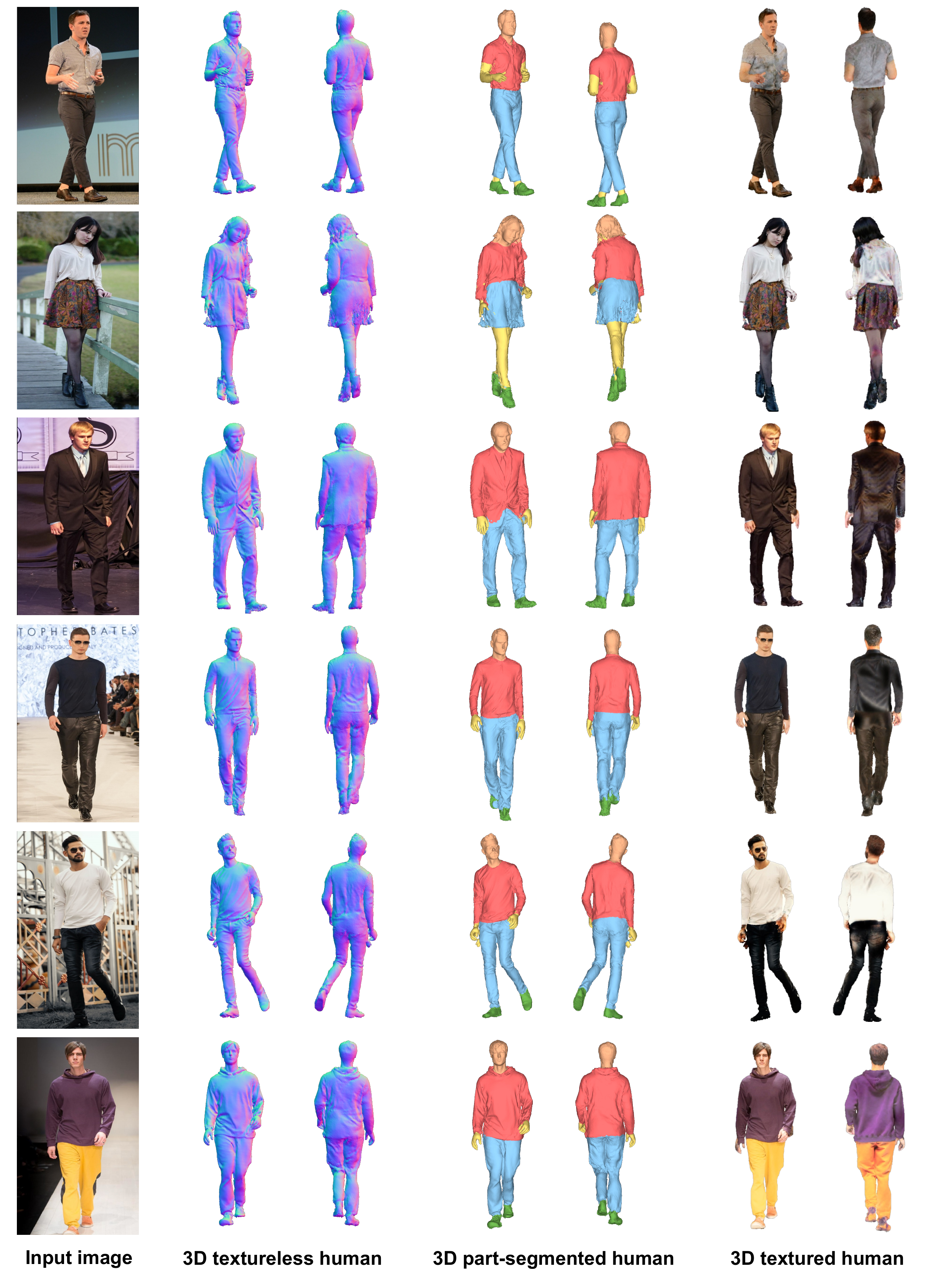}
  \vspace{+0.2em}
  \caption{\textbf{More qualitative results of \ourmethod~on in-the-wild images of SHHQ~\cite{fu2022stylegan}.}
  }
  \vspace{+1.0em}
  \label{fig:more_results_2}
\end{figure*}

\begin{figure*}[t]
  \centering
  \includegraphics[width=0.88\linewidth]{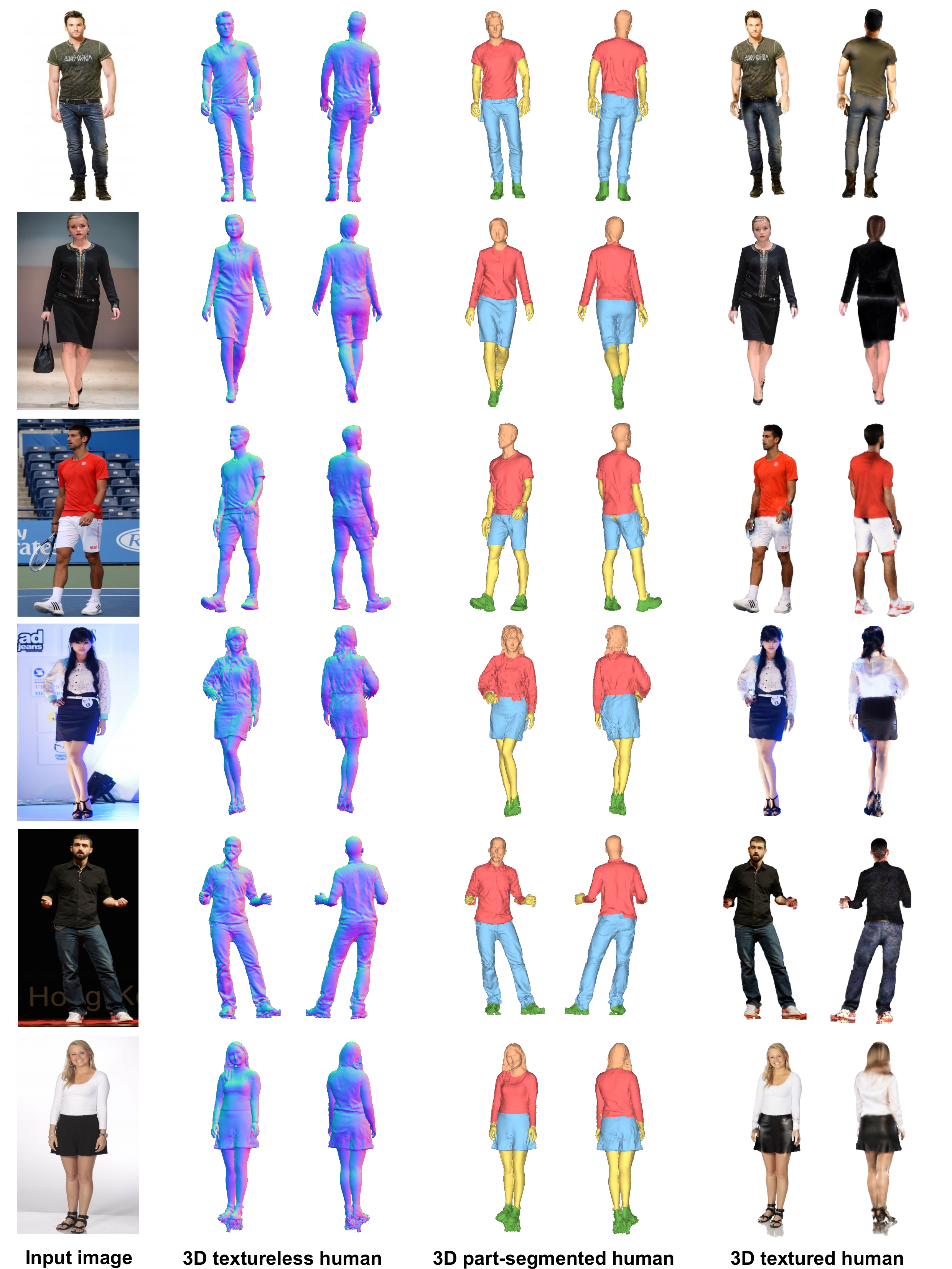}
  \vspace{+0.2em}
  \caption{\textbf{More qualitative results of \ourmethod~on in-the-wild images of SHHQ~\cite{fu2022stylegan}.}
  }
  \vspace{+1.0em}
  \label{fig:more_results_3}
\end{figure*}

We provide more qualitative results of our~\ourmethod~on THuman2.1~\cite{tao2021function4d} and SHHQ~\cite{fu2022stylegan}.
\cref{fig:more_comparison} demonstrates that~\ourmethod~achieves significantly superior texture reconstruction compared to previous 3D human reconstruction methods, demonstrating better part alignment and visual fidelity. 
Figs.~\ref{fig:more_results_1},~\ref{fig:more_results_2}, and~\ref{fig:more_results_3} illustrate that our framework effectively handles in-the-wild scenarios.
\cref{tab:4d_dress} shows that our~\ourmethod~also outperforms the existing reconstruction methods on 4D-DRESS~\cite{wang20244d}, a dataset that has accurate 3D part labels.
For evaluation on 4D-DRESS, we uniformly sample 16 GTs from its test set.

\section{Limitations and future works}
\label{sec:limitation}
\noindent\textbf{Unseen cloth types.}
\cref{fig:suppl_limitation} illustrates the failure cases of our framework when reconstructing unseen cloth types (\textit{e.g.}, dresses) that are not included in the training dataset.
Our training set is labeled based on the pre-defined part categories of Sapiens~\cite{khirodkar2024sapiens}, which is limited to classifying clothes into two types, upper- and lower-clothes.
However, this categorization does not include dresses or multi-layered outfits, resulting in segmentation failures for such cases.
These segmentation failures lead to incorrect human texturing.
We aim to extend our framework to handle various cloth styles by enriching the training data with more diverse cloth samples.

\noindent\textbf{Fine-grained part segmentation.}
Our framework segments the human into $n=5$ part categories, primarily focusing on broad regions.
However, real-world human appearance includes detailed human body parts (\textit{e.g.}, hair and eyes) with various accessories (\textit{e.g.}, hat, glasses, and watch), which are not explicitly segmented in our framework.
A potential future direction is to incorporate fine-grained part segmentation to improve texture reconstruction by accurately distinguishing these intricate elements.

\clearpage

\clearpage

{
\small
\bibliographystyle{ieeenat_fullname}
\bibliography{egbib}
}

\end{document}